\documentclass[lettersize,journal]{IEEEtran}
\usepackage{amsmath,amsfonts}
\usepackage{algorithmic}
\usepackage{array}
\usepackage[caption=false,font=normalsize,labelfont=sf,textfont=sf]{subfig}
\usepackage{textcomp}
\usepackage{stfloats}
\usepackage{url}
\usepackage{verbatim}
\usepackage{graphicx}

\usepackage{amsmath}
\usepackage{amssymb}
\usepackage{booktabs}
\usepackage{dsfont}
\usepackage{adjustbox}
\usepackage{color}
\usepackage{colortbl}
\usepackage{xcolor}
\usepackage{array}

\usepackage{algorithm}
\usepackage{algorithmic}

\usepackage[colorlinks=true, linkcolor=blue, citecolor=blue, urlcolor=cyan]{hyperref}

\definecolor{mygray}{gray}{.9}
\usepackage{multirow}
\def\BibTeX{{\rm B\kern-.05em{\sc i\kern-.025em b}\kern-.08em
    T\kern-.1667em\lower.7ex\hbox{E}\kern-.125emX}}
\usepackage{balance}
\begin{document}
\title{Not Every Patch is Needed: Towards a More Efficient and Effective Backbone for Video-based Person Re-identification}
\author{Lanyun Zhu, Tianrun Chen, Deyi Ji, Jieping Ye, \IEEEmembership{Fellow, IEEE}, and Jun Liu \thanks{Lanyun Zhu is with Information Systems Technology
and Design Pillar, Singapore University of Technology and Design, Singapore, 487372 (e-mail: lanyun\_zhu@mymail.sutd.edu.sg)} \thanks{Tianrun Chen is with College of Computer Science and Technology,
Zhejiang University, China, 310027 (email: tianrun.chen@zju.edu.cn)} \thanks{Deyi Ji and Jieping Ye are with Alibaba Group, China, 310023 (email: jideyi.jdy@alibaba-inc.com, yejieping.ye@alibaba-inc.com)} \thanks{Jun Liu is with School of Computing and Communications, Lancaster University, UK (e-mail: j.liu81@lancaster.ac.uk)}\thanks{This work is supported by the National Research Foundation, Singapore under its AI Singapore Programme (AISG Award No: AISG2-PhD-2021-08-006, and AISG-100E-2023-121).}
\thanks{Corresponding author: Jun Liu.}}

\markboth{Journal of \LaTeX\ Class Files,~Vol.~18, No.~9, September~2020}%
{Not Every Patch is Needed: Towards a More Efficient and Effective Backbone for Video-based Person Re-identification}

\maketitle

\begin{abstract}
This paper proposes a new effective and efficient plug-and-play backbone for video-based person re-identification (ReID). Conventional video-based ReID methods typically use CNN or transformer backbones to extract deep features for every position in every sampled video frame. Here, we argue that this exhaustive feature extraction could be unnecessary, since we find that different frames in a ReID video often exhibit small differences and contain many similar regions due to the relatively slight movements of human beings. Inspired by this, a more selective, efficient paradigm is explored in this paper. Specifically, we introduce a patch selection mechanism to reduce computational cost by choosing only the crucial and non-repetitive patches for feature extraction. Additionally, we present a novel network structure that generates and utilizes pseudo frame global context to address the issue of incomplete views resulting from sparse inputs. By incorporating these new designs, our backbone can achieve both high performance and low computational cost. Extensive experiments on multiple datasets show that our approach reduces the computational cost by 74\% compared to ViT-B and 28\% compared to ResNet50, while the accuracy is on par with ViT-B and outperforms ResNet50 significantly.
\end{abstract}


\section{Introduction}
\IEEEPARstart{P}{erson} ReID \cite{zheng2016person,li2021diverse, chen2019abd, zheng2019joint, rao2021counterfactual, zhou2019omni, huang2023reasoning, huang2023enhancing, wu2023learning} is an important research topic that has been actively investigated for decades. In recent years, for the industry and research community, video-based ReID \cite{li2019global, zhang2020multi, hou2021bicnet, liu2021watching, aich2021spatio, eom2021video, chen2020temporal, lin2022learning} is catching more attention with its advantageous properties like containing additional temporal cues and more comprehensive appearance information compared to image-based methods. To unleash the potential of video-based ReID, previous works have explored various techniques to extract effective features from videos. For example, \cite{li2019multi, gu2020appearance, zhang2017learning} exploit temporal information to enhance person representations; \cite{yang2020spatial, wu2020adaptive, yan2020learning} utilize spatial-temporal correlations based on graph neural networks. The recent success of vision transformers in computer vision has propelled the development in this domain, opening up possibilities for further improvements towards new performance heights.

Despite their success, these existing video-based ReID methods still suffer from a common problem -- huge computational burden. Most of the above-mentioned approaches process the video by passing each of the sampled frames through a backbone network to get a frame feature and then merge different frames to form a representation of the entire video. Although this is a typical way of processing video data for deep networks, it will cause the computation cost to rise linearly with the number of sampled frames in a video clip, thus resulting in a huge computational load. This challenge is even more severe when using the latest transformer-based backbone networks like ViT, in which videos are processed into multiple patches and a huge computational cost is required to correlate them. Consequently, although with better performance than CNN-based methods, it becomes almost impossible to use these transformer backbones in practical applications for ReID, especially in many real-world cases where real-time processing is required.

This work aims to address this challenge for computation by providing an innovative alternative path. We begin by asking a more fundamental question:
\textit{\textbf{Is it really necessary to put so many computational resources just to process every region in a frame to fully exploit the information contained in a video?}} After analyzing videos used for person ReID in detail, we derive a `No' answer to this question. Specifically, we found two properties in the videos used for ReID that can motivate our computational-efficient solutions. Firstly, the region of interest (the human) can remain in a relatively consistent position in all frames, since the videos are often pre-processed and cropped around target persons. Secondly, the variation caused by human motion is also minimal because intense movements like jumping or striking rarely appear in ReID videos. Considering the two properties, we find that ReID videos typically exhibit minimal temporal variations and have a significant number of similar regions existing in different frames. Intuitively, the repetitive feature extraction for these similar regions is redundant and unnecessary. 

Based on the above insights, in this paper, we present a novel and plug-and-play backbone framework for video ReID that leverages the performance advantage of ViT but in a more efficient way with less computational burden. Our proposed framework achieves excellent performance while maintaining reasonable computation costs, achieving competitive results as ViT but only requiring even lower computations than existing CNN-based backbones. The balance of accuracy and efficiency is achieved by our central idea of pruning redundant patches within each frame and selecting only the crucial patches as transformer input for deep feature extraction. By doing so, we effectively reduce both computation requirement and feature redundancy, thus resulting in a combination of ultra-performance and efficiency. Specifically, we first propose a novel mechanism for patch selection. Differing from previous methods of token pruning \cite{chen2023diffrate,wang2022efficient} in other tasks, our patch selection is carefully designed with a frame-progressive strategy and multi-level selection-determinate features that are tailored for the ReID task. By using it, crucial patches that are useful for ReID can be automatically selected while redundant ones that repeat across frames or exhibit unnecessary properties can be pruned. Next, we design a new feature extractor, namely patch-sparse transformer (PSFormer) to better utilize the selected patches. The motivation of this new extractor is that we find conventional feature extractors suffer from a loss of global information when handling the sparse input from each frame and thus yielding suboptimal results. Therefore, we propose a self-adaptive dynamic routing mechanism to generate pseudo frame global context in our PSFormer, making it capable of effectively handling sparse inputs. 

In terms of extracting cross-frame temporal correlations, which is important in video-based ReID, we also propose a novel and efficient approach by using Group of Pictures (GOP). We noted that earlier works explored methods like RNNs \cite{zhang2017learning, mclaughlin2016recurrent}, optical flows \cite{chen2018video, liu2017video}, GNNs \cite{yang2020spatial, yan2020learning, wu2020adaptive}, or events \cite{cao2023event} to extract these correlations, but they typically require high computation costs or additional sensors, which is contradictory to our high-efficiency goal. We hereby take an alternative path by pioneering introducing the use of GOP to represent temporal correlations, which is a data structure encoded in compressed videos. For real-world ReID systems, video compression is a necessary step for data transmission between camera terminals and servers. Therefore, the GOP can be directly obtained from the data received by the server, without additional computational costs. Furthermore, the GOP stores information patch-by-patch, which is inherently aligned with the data format of the transformers. Therefore, we use GOP in both the patch selection mechanism and PSFormer in our approach, resulting in a very efficient ReID framework.

We conduct extensive experiments on multiple datasets and the results show the high effectiveness and efficiency of our method. Specifically, our backbone reduces the computational cost by 74\% compared to ViT-B and 28\% compared to ResNet50, while the results are on par with ViT-B and outperform ResNet50 significantly. Our method is also tested as a plug-and-play backbone for existing methods, with excellent results demonstrating its high generality. In summary, the main contributions of this work are as follows: 
\begin{itemize}
    \item We propose a patch selection mechanism to avoid redundant feature extraction on repetitive or unimportant regions, thereby reducing computational overhead.
    \item We introduce a patch-sparse transformer to address the issue of context loss when dealing with sparse inputs, thereby enhancing the model's effectiveness.
    \item By incorporating the patch selection mechanism and patch-sparse transformer, we get a novel backbone that can simultaneously achieve high effectiveness and efficiency in video ReID, as demonstrated by our extensive experiments on multiple datasets. 
\end{itemize}

\section{Related Work}
\subsection{Video-based Person ReID}
Video-based person ReID aims to extract comprehensive person representations from videos to facilitate accurate identification. Previous methods have explored mechanisms like RNNs \cite{zhang2017learning, mclaughlin2016recurrent}, 3D CNNs \cite{li2019multi, gu2020appearance, liu2021spatial}, optical flows \cite{chen2018video, liu2017video}, events \cite{cao2023event}, attentions \cite{yang2024stfe, wang2021pyramid}, and GNNs \cite{yang2020spatial, yan2020learning, wu2020adaptive} to capture cross-frame temporal information. For example, \cite{yang2024stfe} proposes a feature space projection module to preserve more frequency information and avoid feature fragmentation, and introduces a global low-frequency enhancement module to capture global low-frequency features and establish spatial-temporal relationships across video sequences; \cite{li2019multi} proposes a two-stream convolution network built from 3D CNNs to capture both spatial and temporal information to help enhance video ReID; \cite{chen2018video} improves video ReID by introducing a competitive snippet-similarity aggregation method with the help of optical flow information; \cite{cao2023event} utilizes information extracted from event cameras to address the inevitable degradation issues such as motion blur that can adversely affect video ReID. \cite{liu2023frequency} regards the complex temporal features of video ReID as a kind of signal and converts it into frequency domain to extract the useful spectrum cues. \cite{yu2023tf} introduces a temporal memory diffusion module after the Clip model to extract temporal information by leveraging the sequence-level relations. However, these methods typically require building additional modules that incur extra computational costs to extract temporal cues. In contrast, we propose the first method that uses compressed videos to extract temporal features, which is more efficient and does not require extra sensors or computations like the above-mentioned previous methods. Other methods \cite{fu2019sta, subramaniam2019co, zhao2019attribute, zhou2017see} use temporal attention to determine the importance of each frame. For instance, \cite{zhao2019attribute} uses an attribute-driven method to disentangle features and re-weight frames, \cite{wu2022temporal} selects the most discriminative frames automatically. These approaches, however, focus mainly on frame-level importance, neglecting the more fine-grained region-level importance. In \cite{bai2022salient} and \cite{hou2020temporal}, later frames are encouraged to focus on different regions from the previous frames, but it still requires substantial computational resources to extract deep features from each frame. Alternatively, we propose a pre-network patch selection mechanism that selects crucial patches from a video before processing them through a deep network, making the feature extraction process more efficient and effective. To ensure the effectiveness of our network, we propose warping the initial frame's features to generate features for subsequent frames. While \cite{hou2019vrstc} generates pseudo features as well, its purposes and methods differ from ours significantly. \cite{hou2019vrstc} uses the vanilla attention to generate the occlusion part features, while we propose a GOP-guided warping method to generate frame global features. Moreover, we present a self-adaptive strategy for achieving both low computation and high effectiveness, whereas the attention mechanism in \cite{hou2019vrstc} incurs higher computation costs. Another work related to our method is \cite{zhang2024magic}, which also uses a patch selection strategy in ReID to focus on the foreground area where the target person is located. However, it employs a completely different mechanism from ours. Specifically, in \cite{zhang2024magic}, deep network (ViT) features are first extracted for all patches in the entire image, and then key patches are selected based on these ViT features. As a result, each patch still needs to go through the full deep ViT for feature extraction, which incurs a significant computational cost. Our method addresses this issue by proposing a before-deep-network approach where key patches are selected prior to passing through the deep network for feature extraction, thus greatly reducing computational demands. Additionally, \cite{zhang2024magic} only focuses on foreground-background separation, while our method further avoids redundant extraction of similar patches across different frames. With these novel designs, our method can be more efficient.

\subsection{Efficient Vision Transformers and Token Pruning/Selection}
In recent years, networks based on the transformer architecture have achieved tremendous success across various fields in computer vision \cite{carion2020end, zhu2020deformable, cheng2022masked, xie2021segformer, zheng20213d, he2021transreid, zhao2024structure, zhu2024llafs, chen2023sam, zhu2024ibd, chen2024sam2, ji2024discrete, chen2024reasoning3d}. However, the higher computational costs compared to traditional CNNs \cite{he2016deep, zhu2021learning, zhu2023continual, zang2025resmatch, zhu2024addressing, ji2024discrete, liu2021label, chen2023deep3dsketch, chen2023reality3dsketch} have limited the application of these transformer methods in practical scenarios. To address this issue, researchers are actively seeking more efficient vision transformer approaches to reduce computational expenses. Some methods \cite{yu2022metaformer, huang2022orthogonal, han2023flatten, you2024shiftaddvit, you2023castling, li2022efficientformer} propose compact architectures to reduce the computational cost of self-attention. For example, \cite{huang2022orthogonal} introduces an orthogonal self-attention mechanism that reduces the computational load while retaining more detailed information. Other approaches \cite{yuan2022ptq4vit, liu2021post, ding2022towards, liu2023noisyquant} utilize quantization techniques to minimize computational overhead. For instance, \cite{liu2021post} introduces a quantization approach specifically designed for vision transformers based on mixed-precision weights. 
Meanwhile, some methods \cite{touvron2021training, wu2022tinyvit, zhang2022minivit, lin2023supervised} employ distillation techniques to transfer the capabilities of large vision transformers to smaller transformers that require less computation.

Recently, the spatial redundancy in nature images has motivated some studies \cite{tang2022patch, rao2021dynamicvit, meng2022adavit, wei2023joint, ma2022ppt, liang2022not, kong2022spvit} to explore another kind of methods that discard nonessential tokens to improve processing speed. These techniques, known as token pruning or token selection, have been extended to temporal networks for efficient video processing \cite{wang2022efficient, li2023svitt, gao2023sparseformer, dutson2023eventful}. For example, \cite{tang2022patch} introduces a patch slimming method that discards useless patches in a top-down manner, with a novel mechanism that applies the crucial patches in a network's higher layers to guide the selection of patches in the shallower layers. \cite{wang2022efficient} measures the importance of every region and chooses those with top scores to be used for downstream processing. \cite{liang2021evit} proposes an automatic approach to prune unnecessary tokens that are semantically meaningless or distractive image backgrounds. \cite{zhu2023learning} employs a detection module and scoring function to select only the class-discriminative regions for the computationally expensive texture feature extraction. \cite{li2023svitt} introduces a sparse video-text architecture with two forms of sparsity including edge sparsity to reduce the inter-token query-key communications and node sparsity to discard less-crucial tokens. \cite{gao2023sparseformer} proposes a novel network that imitates human’s sparse visual recognition in an end-to-end manner by representing images using only a very limited number of tokens in the latent space. Despite some success, these methods still exhibit two significant drawbacks. First, the pruning or selection mechanisms of these methods are not specifically designed for video-based person ReID, so directly applying them to this task may not guarantee task-optimal results. Second, most of these methods only focus on selecting sparse tokens, without further considering how to enhance the network's effectiveness with just sparse inputs. Unlike these methods, we propose a novel framework that resolves these issues, making the video ReID task more efficient and effective. Performance comparisons with these token pruning methods are presented in Sec. \ref{comparision_text}.

\section{Preliminaries} \label{pre}
\subsection{Group of Pictures}
Group of Pictures (GOP) is a data structure encoded in compressed videos \cite{wiegand2003overview, lin2009versatile, lu2022learning, zhu2019high, fan2021motion} to reduce the video storage cost. In a GOP, a video's first and subsequent frames are represented differently. The first frame, referred to as the \textbf{I-frame}, is represented by a fully encoded image $I\in \mathbb{R}^{3\times H\times W}$. Each of the subsequent frames, known as the \textbf{P-frame}, is represented by a motion vector $M_{t}\in \mathbb{R}^{2\times \frac{H}{16}\times \frac{W}{16}}$ combined with a residual map $R_{t}\in \mathbb{R}^{3\times H\times W}$. The $(i,j)$-th pixel $M_{t}^{i,j}$ on $M_{t}$ represents a coordinate, which indicates the displacement of the most similar patch in I-frame to the $(i,j)$-th patch in the $t$-th P-frame. $R_{t}^{:, 16(i-1):16i, 16(j-1):16j}$ represents residual errors by predicting the P-frame patch based on its I-frame displacement.

\subsection{Patch-wise Representation}
In transformer-based vision models like ViT, we generally represent the image patch-wise for sequential processing. Specifically, given an image with size of $\mathbb{R}^{3\times H\times W}$, we first divide it equally into $H/16\times W/16$ patches, each sized $16\times 16$, and then flatten it into the shape $\mathbb{R}^{N\times 768}$, where $N=H/16\times W/16$, $768=3\times 16\times 16$ is the dimension of the flattened vector for each patch. To facilitate further operations, we reshape all maps within a GOP into a patch-based representation in this manner, getting $I$, $M_{t}$ and $R_{t}$ with the shape of $\mathbb{R}^{N\times 768}$, $\mathbb{R}^{2\times N}$ and $\mathbb{R}^{N\times 768}$ respectively. Each $M_{t}^{i}\in \mathbb{R}^{2}$ in $M_{t}$ represents a 2D spatial coordinate $(h, w)$, which we further convert into a single index $v$ to correspond to the flattened coordinate, using the formula: $v = W/16 * h + w$. In this way, we get $M_{t}\in \mathbb{R}^{N}$.

\subsection{Spectral Decomposition} \label{text_spectral}
Spectral decomposition is a method proposed by \cite{melas2022deep} that can locate the prominent object in an image. Given an input feature $F \in \mathbb{R}^{H\times W}$, this method firsts computes an affinity matrix $A\in \mathbb{R}^{N\times N}$ \footnote{$N=H\times W$} through $A^{i,j}=F^{i}.(F^{j})^{\rm T}$, where $F^{i}$ refers to the $i$-th pixel on $F$, and then calculates the normalized Laplacian matrix $L$ from $A$ by using the formula $L = D^{-1/2}(D - A)D^{-1/2}$, where $D$ is a diagonal matrix with $D^{i,j}=\sum_j A^{i,j}$. The eigenvectors of $L$ are then computed and denoted as $\{y_{0}, y_{1}, ... y_{N-1}\}$, which are sorted according to their corresponding eigenvalues $\lambda_{0}\leq \lambda_{1}...\leq\lambda_{N-1}$ in the ascending order. Based on the empirical observations in \cite{melas2022deep}, the eigenvector $y_{1}\in \mathbb{R}^{N}$ associated with the smallest nonzero eigenvalue $\lambda_{1}$ typically represents the most prominent object in a scene. In this work, we leverage this characteristic to develop a novel patch selection mechanism.

\begin{figure*}
    \centering
    \includegraphics[width=1\linewidth]{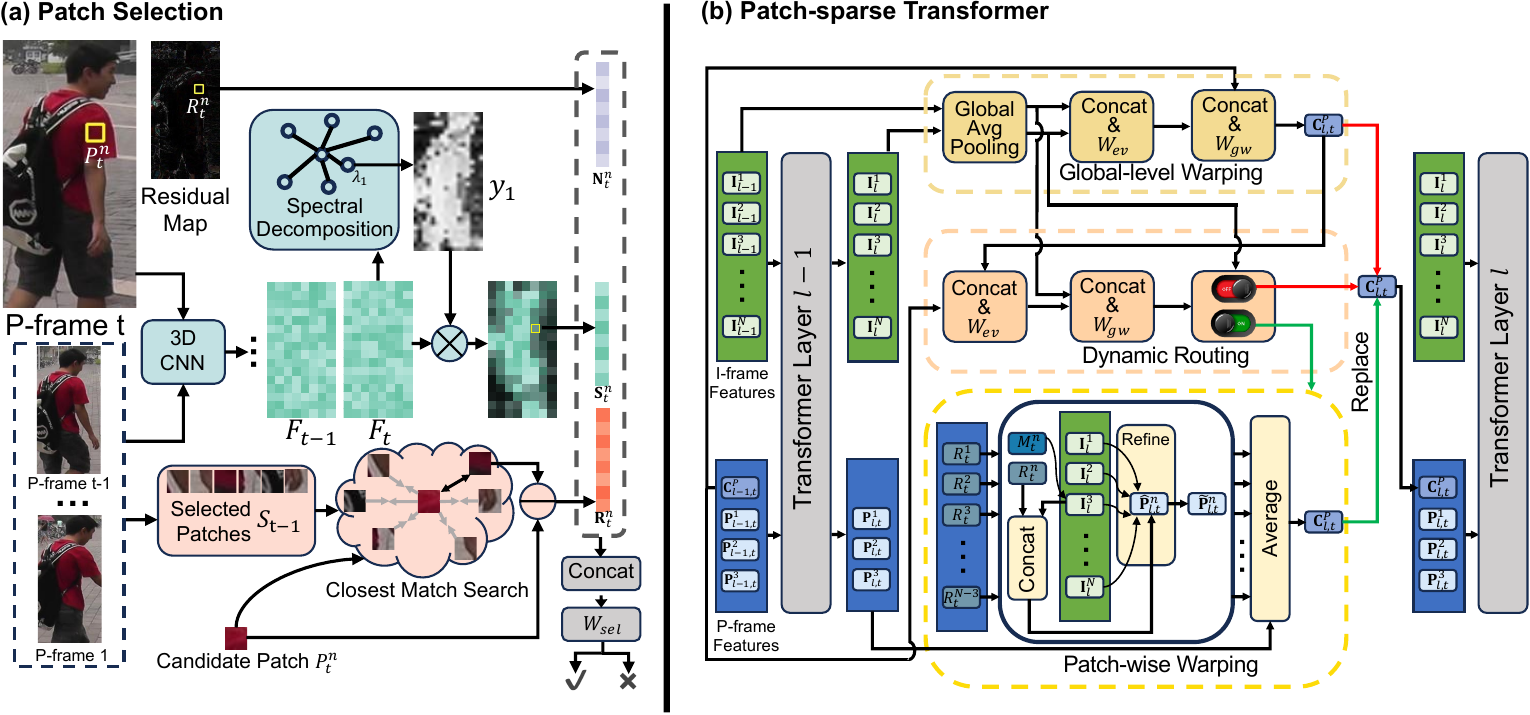}
    \caption{\textbf{Overview of our framework}, including two components: (a) Patch Selection and (b) Patch-sparse Transformer. Note that, for simplicity of illustration, this figure only presents one I-frame with one P-frame in the patch-sparse transformer.}
    \label{main_fig}
\end{figure*}

\section{Proposed Method}
\subsection{Overview} Following ViT \cite{dosovitskiy2020image}, given a video clip $V\in \mathbb{R}^{3\times H\times W\times T}$, our framework first transfers each frame to a patch-wise representation with size of $N\times 768$. Next, instead of extracting person features from all patches in all frames like previous transformer methods, we propose to use only a few patches selected from a video to be processed by the deep network. The model computation and feature redundancy can be thus greatly reduced. Specifically, we first propose a novel mechanism (Sec.\ref{select}) to automatically select patches. The selected patches are then fed into our proposed patch-sparse transformer (Sec.\ref{network}) to extract fine-grained person features, which are ultimately used to identify persons. In the following sections, we describe each part of the framework in detail.

\subsection{Patch Selection} \label{select}
The effectiveness of our framework is guaranteed by carefully selecting the best patches for feature extraction. Here, we utilize the GOP introduced in Sec.\ref{pre} that does not require additional sensors or computations to help this selection process. Specifically, we treat the I-frame within a GOP as the primary frame and use all patches within it as model inputs. We find retaining the complete I-frame is essential because it allows the extraction of important global features such as body shape, which is critical for the ReID task. Conversely, the remaining P-frames are selectively processed as subsidiary frames, from which only a subset of patches are selected for deep feature extraction. 

To effectively and automatically select patches in the P-frames, we propose a new mechanism based on a scoring function as shown in Figure \ref{main_fig} (a) and Algorithm \ref{alg_1}. Specifically, we denote the $n$-th patch in the $t$-th P-frame by $P_{t}^{n}$. For each $P_{t}^{n}$, we capture a selection-determinate feature for it, which is then inputted into a 3-layer MLP $W_{sel}$ to generate a score. This score finally determines whether a patch should be selected or not. As the MLP input, the selection-determinate feature is the key to making this mechanism work. In the following subsection, we introduce this feature in detail.

\subsubsection{Selection-determinate Feature}\label{text_selection_feature} To guarantee an effective decision-making process with comprehensive considerations, we introduce two criteria as candidates for the selection-determinate feature: Patch Novelty $\mathbf{N}_{t}^{n}$ and Patch Semantics $\mathbf{S}_{t}^{n}$. The motivation for using these two criteria is that we hope the selected patches are novel to the I-frame and not located in background areas, thus avoiding the redundant feature extraction for the cross-frame repetitive and ReID-unrelated regions. Formally, $\mathbf{N}_{t}^{n}$ is represented by the residual map $R_{t}^{n}$ within a GOP, which measures the dissimilarity between the patch and the I-frame. $\mathbf{S}_{t}^{n}$ represents a feature with semantic information related to the selection decision, which is obtained through a two-step process to extract and enhance features. Specifically, in the first step, we pass the input video through a shallow 4-layered 3D CNN to generate a feature map $F_{t} \in \mathbb{R}^{H/16 \times W/16 \times C}$ for each frame, which contains temporal and low-level information. In the second step, we utilize spectral decomposition to enhance $F_{t}$ with saliency features, enabling it to capture foreground-background partitions more effectively. Specifically, we compute the eigenvector $y_1$ associated with the smallest nonzero eigenvalue $\lambda_{1}$ from $F_{t}$ using the spectral decomposition method introduced in Sec.\ref{text_spectral}. Through our empirical observations presented in Sec.\ref{text_effect_spectral}, $y_1$ typically captures the region of the target person in a ReID image. Therefore, we reshape $y_{1}$ to $\mathbb{R}^{H/16\times W/16}$ and use it as a target-person indicator to enhance $F_{t}$ by computing $\mathbf{S}_{t}=F_{t}\cdot y_{1}$. In this way, the resulting $\mathbf{S}_{t}$ incorporates both patch semantics and foreground awareness, so it can be used to help $W_{sel}$ to select better patches with crucial semantic information and within the target person.

\subsubsection{Progressive Selection} Using $W_{sel}$ in conjunction with input factors $\mathbf{N}$ and $\mathbf{S}$, we can calculate patch scores for P-frames and select patches accordingly. The way of calculating scores for different P-frames is the following problem that needs to be carefully considered. Instead of calculating scores for all P-frames at once, we propose to use a progressive approach where patches are sequentially selected from the first frame to the last. This is because if we calculate scores in parallel, redundancy would occur when two similar patches originating from different P-frames are simultaneously chosen due to their closely matched input factors. Therefore, in our progressive approach, we construct a set $\mathcal{S}_{t-1}$ containing all patches selected from the 1st to the current P-frames once the patch selection is completed in the $t-1$-th P-frame. When calculating the score for each patch $P_{t}^{n}$ during the selection process of the subsequent frame $t$, we conduct a search for its closest match within $\mathcal{S}_{t-1}$\footnote{The patch similarity is measured using the same method employed in the H.264 video compression standard.} and then calculate the pixel-wise residual $\mathbf{R}_{t}^{n}$ between $P_{t}^{n}$ and this matched patch. This residual information is incorporated as the third input to $W_{sel}$ for scoring. By introducing $\mathbf{R}_{t}^{n}$ into the scoring process, we make $W_{sel}$ to have the awareness and adaptability to avoid selecting patches that closely resemble those already chosen, thereby effectively addressing the aforementioned redundancy issue by avoiding selecting similar patches repetitively.

\subsubsection{Differentiable Selection}Eventually, we concatenate $\mathbf{N}_{t}^{n}$, $\mathbf{S}_{t}^{n}$ and $\mathbf{R}_{t}^{n}$, which are inputted into $W_{sel}$ to generate the score $s_{t}^{n}$. $s_{t}^{n}$ is then used to make the selection decision. Although some past ReID works have explored methods to select key regions for emphasis, they either use a soft attention mechanism by assigning different importance weights to different patches and performing a weighted sum \cite{wang2021pyramid}, which is differentiable but cannot achieve a hard decision, thus unable to reduce computational load by excluding unnecessary patches; or they directly use a topK operation for patch selection \cite{zhang2024magic}, which can achieve a hard decision but is non-differentiable, therefore preventing the selection network from receiving gradients for updating. To overcome the limitations of these methods, we propose a novel mechanism to achieve differentiable and hard decisions by utilizing the saturating Sigmoid function \cite{kaiser2018fast} for patch selection. Specifically, during the training stage, we explore more decision space randomly by adding a standard Gaussian noise to $s_{t}^{n}$ and getting $\hat{s}_{t}^{n}$. After that, we generate a binary score using $b_{t}^{n} = \mathds{1}\left(\hat{s}_{t}^{n} > 0\right)$, and obtain a differentiable value $d_{t}^{n}$ by:
\begin{equation}
    d_{t}^{n} = {\rm max}\left(0, {\rm min}\left(1,1.2\sigma\left(\hat{s}_{t}^{n}\right)-0.1\right)\right),
\end{equation}
where $\sigma$ refers to the original Sigmoid function. Here, we use the binary score $b_{t}^{n}$ for selection, i.e, selecting patch $P_{t}^{n}$ if $b_{t}^{n} = 1$, and utilize the differentiable $d_{t}^{n}$ to approximate gradients in back-propagation. Specifically, we use the gradient of $d^{n}_{t}$ with respect to $\hat{s}^{n}_{t}$ to serve as an approximation for the gradients required to update the parameters associated with the discrete gate $b^{n}_{t}$. The method for this gradient approximation can be implemented in PyTorch as follows: $b^{n}_{t} = b^{n}_{t} + d^{n}_{t} - d^{n}_{t}.{\rm detach()}$. Furthermore, during the inference phase, the process of sampling Gaussian noise is omitted. Instead, the discrete output is derived directly from the gate's original score and used for making decisions, i.e., selecting a patch if $s^{n}_{t} > 0$. In this way, the selection operation for $P_{t}^{n}$ is completed, with a detailed description of the overall procedures shown in Algorithm \ref{alg_1}. By using this method, we achieve effective selection that is both differentiable and capable of making hard decisions, thereby reducing computational load and enabling end-to-end training to simplify optimization.

\begin{algorithm}[t]
\caption{Patch Selection Algorithm}
\label{alg_1}
\begin{algorithmic}[1]
\STATE{\textbf{Input:} video clip $V\in \mathbb{R}^{3\times H\times W\times T}$ with $T$ frames, including one I-frame $I\in \mathbb{R}^{3\times H\times W}$ and $T-1$ P-frames $\{P_{t}\}_{t=1}^{T-1}\in \mathbb{R}^{3\times H\times W}$; residual map $R_{t}\in \mathbb{R}^{3\times H\times W}$ in the GOP.}
\STATE{Reshape $I$ to $\mathbb{R}^{N\times 768}$, $P_{t}$ to $\mathbb{R}^{N\times 768}$, $R_{t}$ to $\mathbb{R}^{N\times 768}$.}
\STATE{Pass the video $V$ through a shallow 3D CNN, generating feature map $F_{t} \in \mathbb{R}^{H/16 \times W/16 \times C}$ for each P-frame $P_{t}$. }
\STATE{Initialize $\mathcal{S}_{0}$ as an empty set.}
\FOR{$t$ \textbf{in} $1,...,T-1$}
\STATE{\textit{\textbf{1. Obtain Patch Novelty}} $\textbf{N}_{t}=\{\textbf{N}_{t}^{n}\}_{n=1}^{N}$:}
\STATE{$\textbf{N}_{t}^{n}\gets R_{t}^{n}$}.
\STATE{\textit{\textbf{2. Obtain Patch Semantics}} $\textbf{S}_{t}=\{\textbf{S}_{t}^{n}\}_{n=1}^{N}$:} 
\STATE{Get $A\in \mathbb{R}^{N\times N}$ through $A^{i,j}=F_{t}^{i}.(F_{t}^{j})^{\rm T}$ ($N=H/16\times W/16$);}
\STATE{Get $D\in \mathbb{R}^{N\times N}$ through $D^{i,j}=\sum_j A^{i,j}$;}
\STATE{Get $L\in \mathbb{R}^{N\times N}$ through $L = D^{-1/2}(D - A)D^{-1/2}$; }
\STATE{Compute the eigenvectors $\{y_{0}, y_{1}, ... y_{N-1}\}$ of $L$ sorted according to their corresponding eigenvalues $\lambda_{0}\leq \lambda_{1}...\leq\lambda_{N-1}$ in the ascending order;}
\STATE{Reshape $y_{1}$ to $\mathbb{R}^{H/16\times W/16}$, compute $\mathbf{S}_{t}=F_{t}\cdot y_{1}$. }
\STATE{\textit{\textbf{3. Obtain}} $\textbf{R}_{t}=\{\textbf{R}_{t}^{n}\}_{n=1}^{N}$ \textit{\textbf{{for progressive selection:}}}}
\STATE{Search the most similar patch $p$ in $\mathcal{S}_{t-1}$ for $P_{t}^{n}$;}
\STATE{Compute pixel-wise residual $\mathbf{R}_{t}^{n}$ between $P_{t}^{n}$ and $p$.}
\STATE{\textit{\textbf{4. Selection Decision for}} $\{P_{t}^{n}\}_{n=1}^{N}$:}
\STATE{Get $s_{t}^{n}$ through $s_{t}^{n}=W_{sel}(\mathbf{N}_{t}^{n}||\mathbf{S}_{t}^{n}|| \mathbf{R}_{t}^{n})$;}
\STATE{Select $P_{t}^{n}$ if $s_{t}^{n}>0$.}
\STATE{\textit{\textbf{5. Update}} $\mathcal{S}$:}
\STATE{Get $\mathcal{S}_{t}$ by appending all selected patches $\{P_{t}^{n}\}_{n=1}^{\hat{N}_{t}}$ from $P_{t}$ to $\mathcal{S}_{t-1}$.}
\ENDFOR
\STATE{\textbf{Return:} All patches $\{I^{n}\}_{n=1}^{N}$ in I-frame and the selected patches $\{\{P_{t}^{n}\}_{n=1}^{\hat{N}_{t}}\}_{t=1}^{T-1}$ from P-frames.}
\end{algorithmic}
\end{algorithm}

\subsection{Patch-sparse Transformer} \label{network}
\subsubsection{Motivation}The previously mentioned method enables key patches to be selected from each P-frame. These patches can be input into a regular transformer like ViT for feature extraction with reduced computational cost. However, it is observed that when using the vanilla ViT as a feature extractor, employing these carefully selected patches as sparse inputs still results in a significant performance gap compared to using the entire video input. This gap could be attributed to the loss of global-aware information during the feature extraction process, which requires a non-local perspective so that cannot be captured when solely analyzing a limited number of patches. For example, without the complete view of a P-frame, it is difficult to identify whether a pattern in a selected patch is located on clothing or pants, which is a critical cue for person ReID. To address this issue, an intuitive and effective approach is to introduce global context so that the model can perceive global properties more effectively. However, trivial methods of obtaining context, such as global average pooling, are not directly applicable to our method since most P-frame positions do not have extracted features. As a result, context-based enhancement becomes a very challenging task in our framework. Fortunately, we observe that P-frames typically exhibit high similarities to the I-frame in a ReID video. This inspires us to propose a novel method that harnesses the I-frame features to generate pseudo global context $\mathbf{C}_{l,t}^{P}$ for each P-frame $P_{t}$ in each transformer layer $l$ and then use it to enhance global perception of the network. By doing so, as shown in Figure \ref{main_fig} (b), a novel patch-sparse transformer (PSFormer) can be developed to effectively handle sparse inputs (Sec.\ref{text_trans}). To simultaneously balance effectiveness and efficiency, we propose two complementary methods for generating pseudo context, namely patch-wise warping and global-level warping as follows:

 \noindent \textbf{ (a) Patch-wise Warping} generates pseudo context by first producing a pseudo feature for each unselected patch in P-frames. We notice that the GOP framework introduced in Sec.\ref{pre} can provide a perfect tool for this warping process. Specifically, in a GOP, the motion vector $M$ represents the pairwise alignment between each patch of a P-frame and its displacement patch in the I-frame. According to the construction mechanism of GOP, these two patches are typically similar, with only minor differences being recorded in the residual map ${\rm R}$. Taking advantage of this characteristic, we leverage the features of the displacement to generate a pseudo feature for the P-frame patch by warping using the residual map that reflects the difference information. Formally, in the $l$-th layer of the transformer, we denote the feature for all I-frame patches as $\{\mathbf{I}_{l}^{i}\}_{i=1}^{N}$. And for each unselected patch $P_{t}^{n}$ in the $t$-th P-frame, the transformer feature for its displacement patch is denoted as $\mathbf{I}_{l}^{M_{t}^{n}}$. As shown in the bottom part of Figure \ref{main_fig} (b), we combine $\mathbf{I}_{l}^{M_{t}^{n}}$ with the residual map $R_{t}^{n}$, and derive the pseudo feature $\hat{\mathbf{P}}_{l,t}^{n}$ for $P_{t}^{n}$ through the operation $\hat{\mathbf{P}}_{l,t}^{n} = W_{pw}(\mathbf{I}_{l}^{M_{t}^{n}}||R_{t}^{n})$\footnote{$||$ refers to concatenation.}, where $W_{pw}$ is a 3-layer MLP. In P-frames, the motion vector is predicted at the patch level, so it may not be the exact ground truth at the pixel-level. As a result, the alignment between P-frame patches and their displacements may not be perfect. To alleviate this potential error in alignment, we propose to assign different fusion importance weights to every location in $\mathbf{I}_{l}=\{\mathbf{I}_{l}^{i}\}_{i=1}^{N}$, thus refining $\hat{\mathbf{P}}_{l,t}^{n}$ by aggregating other potentially-related patches in the I-frame. Formally, 
\begin{equation}
    \widetilde{\mathbf{P}}_{l,t}^n{} = {\rm Softmax}\left(\frac{W_{q}(\hat{\mathbf{P}}_{l,t}^{n})W_{k}\left(\mathbf{I}_{l}\right)^T}{\sqrt{d_k}}\right)W_{v}\left({\mathbf{I}_{l}}\right),
\end{equation}
where $W_{q}$, $W_{k}$ and $W_{v}$ are three fully-connected layers respectively aiming at producing the query, key and value features in cross-attention. Through this method, the weight assignment is implemented by measuring the correlation between $\hat{\mathbf{P}}_{l,t}^{n}$ and each $\mathbf{I}_{l}^{i} \in \{\mathbf{I}_{l}^{i}\}_{i=1}^{N}$, and feature aggregation is then completed by using these weights to perform a weighted summation over $\{W_{v}\left(\mathbf{I}_{l}^{i}\right)\}_{i=1}^{N}$. Finally, the frame global context $\mathbf{C}_{l,t}^{P}$ is calculated as the average over all patch features, including the transformer features ${\mathbf{P}}_{l,t}^{n}$ for selected patches and pseudo features $\widetilde{\mathbf{P}}_{l,t}^{n}$ for unselected patches.

\noindent \textbf{(b) Global-level Warping} directly warps the global context from the I-frame. As shown in the top part of Figure \ref{main_fig} (b), we first compute the average feature over all patches $\{\mathbf{I}_{l}^{i}\}_{i=1}^{N}$ in the I-frame, yielding a global context $\mathbf{C}^{I}_{l}$ in the $l$-th layer of the transformer. Similarly, $\mathbf{C}^{I}_{l-1}$ is obtained from the previous layer $l-1$. We concatenate $\mathbf{C}^{I}_{l}$ with $\mathbf{C}^{I}_{l-1}$, and obtain an evolution feature $\mathbf{E}_{l}$ by:
\begin{equation}\label{global_eq_1}
    \mathbf{E}_{l}=W_{ev}(\mathbf{C}^{I}_{l}||\mathbf{C}^{I}_{l-1}),
\end{equation}
where $W_{ev}$ is a 2-layer MLP. Through this process, $\mathbf{E}_{l}$ can capture how the global context of the I-frame evolves from the previous to the current layer in the transformer. Due to the similarity between P-frames and I-frame, their global contexts display similar evolutionary patterns in a network. Therefore, we use $\mathbf{E}_{l}$ as evolution guidance, generating the $t$-th P-frame's pseudo context $\mathbf{C}_{l,t}^{P}$ by warping $\mathbf{C}_{l-1,t}^{P}$ from the previous layer. Formally,
\begin{equation} \label{global_eq_2}
    \mathbf{C}_{l,t}^{P} = W_{gw}(\mathbf{E}_{l} || \mathbf{C}_{l-1,t}^{P}),
\end{equation}
where $W_{gw}$ is a 2-layer MLP. 

\subsubsection{Context Generation with Dynamic Routing}\label{text_dynamic} The aforementioned two warping strategies have advantages and disadvantages. The patch-wise warping can provide better accuracy with its patch-level operation and the refinement mechanism, but it requires a higher computation cost. In contrast, the global-level warping is efficient, but may yield coarse and inaccurate results as it overlooks spatial details and cross-frame differences. Moreover, when using global-level warping, each layer's $\mathbf{C}_{l,t}^{P}$ would be generated and thus inherit errors from the previous layer's $\mathbf{C}_{l-1,t}^{P}$. Consequently, the errors would accumulate in the transformer.

Based on the above discussions, we aim to strike a balance between the efficiency and effectiveness of these two methods, which is also the goal of our framework. We realize this target by proposing a novel dynamic routing mechanism to achieve self-adaptive feature generation, as shown in the middle part of Fig.\ref{main_fig} (b). Specifically, in each transformer layer $l$, we first apply the global-level warping to generate a pseudo context $\mathbf{C}_{l,t}^{P}$, and then deploy an error-conditioned gate to determine whether a more refined version of $\mathbf{C}_{l,t}^{P}$ should be acquired through patch-wise warping. This gate is conditionally opened when the error amount accumulates to a certain extent. To measure the error amount, we employ global-level warping in a `reverse' manner. To be specific, we use the pseudo context $\mathbf{C}_{l, t}^{P}, \mathbf{C}_{l-1, t}^{P}$ of the P-frame from two adjacent layers, generating an I-frame pseudo context $\hat{\mathbf{C}}_{l}^{I}$ for the $l$-th layer by warping from the context ${\mathbf{C}}_{l-1}^{I}$ \footnote{The I-frame context ${\mathbf{C}}_{l}^{I}$ is computed as the average feature of all I-frame patches$\{\mathbf{I}_{l}^{i}\}_{i=1}^{N}$.} of the previous layer $l-1$. Formally, 
 \begin{equation}
 \hat{\mathbf{E}}_{l,t}=W_{ev}\left(\mathbf{C}_{l, t}^{P}||\mathbf{C}_{l-1, t}^{P}\right);\hat{\mathbf{C}}_{l}^{I}=W_{gw}\left(\hat{\mathbf{E}}_{l,t}||\mathbf{C}_{l-1}^{I}\right), 
 \end{equation} 
where $W_{gw}$ and $W_{ev}$ are the networks used in global-level warping (see Eq.\ref{global_eq_1} and \ref{global_eq_2}). Intuitively, increasing the level of errors accumulated in $\mathbf{C}_{l, t}^{P}$ results in a less effective evolution feature $\hat{\mathbf{E}}_{l,t}$, which leads to the coarser feature warping, making the warped $\hat{\mathbf{C}}_{l}^{I}$ from $\mathbf{C}_{l-1}^{I}$ to exhibit a larger distance from the actual context $\mathbf{C}_{l}^{I}$. Therefore, we calculate the cosine distance $c_{l,t}$ between $\hat{\mathbf{C}}_{l}^{I}$ and $\mathbf{C}_{l}^{I}$, and use it as an approximate measurement of the error amount. $c_{l,t}$ is compared with a pre-defined activation threshold $s$. If $c_{l,t}>s$, we activate the gate, performing patch-level warping to obtain a refined pseudo context, which replaces $\mathbf{C}_{l,t}^{P}$. Otherwise, we keep the gate closed, directly using $\mathbf{C}_{l,t}^{P}$ from global-level warping for further operations. By implementing such a design, our network prioritizes using the coarse but computational-efficient global-level warping, while occasionally resorting to the more refined but computational-costly patch-wise warping only when the error accumulation reaches a certain threshold. 

It is worth noting that the hyper-parameter $s$ provides the flexibility to customize the model according to specific requirements. During the inference stage, if higher performance is the focus, the value of $s$ can be reduced. On the other hand, if faster processing speed is desired, the value of $s$ can be increased with faster processing but slightly sacrificed accuracy. Importantly, these adjustments can be made by only modifying the value of $s$, without the need to retrain the network. Experimental results of this dynamic adjustment are shown in Table \ref{ablation_s}.

\begin{algorithm}[t]
\caption{The $l$-th Layer of Patch-sparse Transformer}
\label{alg_2}
\begin{algorithmic}[1]
\STATE{\textbf{Input:} output features from the $l-1$-th layer, including $\{\mathbf{I}_{l}^{i}\}_{i=1}^{N}$ for all I-frame patches, $\{\{\mathbf{P}_{l,t}^{n}\}_{n=1}^{\hat{N}_{t}}\}_{t=1}^{T-1}$ for all selected P-frame patches; GOP; multi-head self attention ${\rm MSA}_{l}$ in the $l$-th layer.}
\STATE{Process I-frame features: $\{\mathbf{I}_{l+1}^{i}\}_{i=1}^{N}={\rm MSA}_{l}(\{\mathbf{I}_{l}^{i}\}_{i=1}^{N})$.}
\FOR{$t$ \textbf{in} $1,...,T-1$}
\STATE{\textit{\textbf{1. Global-level Warping:}}}
\STATE{$\mathbf{C}_{l}^{I}=\frac{1}{N}\sum_{i=1}^{N}\mathbf{I}_{l}^{i}$; $\mathbf{C}_{l-1}^{I}=\frac{1}{N}\sum_{i=1}^{N}\mathbf{I}_{l-1}^{i}$;}
\STATE{$\mathbf{E}_{l}=W_{ev}(\mathbf{C}^{I}_{l}||\mathbf{C}^{I}_{l-1})$; $\mathbf{C}_{l,t}^{P} = W_{gw}(\mathbf{E}_{l} || \mathbf{C}_{l-1,t}^{P})$.}
\STATE{\textit{\textbf{2. Dynamic Routing:}}}
\STATE{$\hat{\mathbf{E}}_{l,t}=W_{ev}(\mathbf{C}_{l, t}^{P}||\mathbf{C}_{l-1, t}^{P}); \hat{\mathbf{C}}_{l}^{I}=W_{gw}(\hat{\mathbf{E}}_{l,t}||\mathbf{C}_{l-1}^{I})$;}
\STATE{$c_{l,t}={\rm Cos}(\hat{\mathbf{C}}_{l}^{I}, \mathbf{C}_{l}^{I})$.}
\IF{$c_{l,t}<s$}
\STATE{$\{\mathbf{P}_{l+1,t}^{n}\}_{n=1}^{\hat{N}_{t}}={\rm MSA}_{l}(\{\mathbf{P}_{l,t}^{n}\}_{n=1}^{\hat{N}_{t}}||\mathbf{C}_{l,t}^{P})$.}
\ELSE
\STATE{\textit{\textbf{3. Patch-wise Warping:}}}
\FOR{$n$ \textbf{in} $1,...,N-\hat{N}_{t}$}
\STATE{Find the I-frame displacement $\mathbf{I}_{l}^{M_{t}^{n}}$ for the $n$-th unselected patch;}
\STATE{$\hat{\mathbf{P}}_{l,t}^{n} = W_{pw}(\mathbf{I}_{l}^{M_{t}^{n}}||R_{t}^{n})$;}
\STATE{$\widetilde{\mathbf{P}}_{l,t}^n{} = {\rm Softmax}\left(\frac{W_{q}(\hat{\mathbf{P}}_{l,t}^{n})W_{k}\left(\mathbf{I}_{l}\right)^T}{\sqrt{d_k}}\right)W_{v}\left({\mathbf{I}_{l}}\right).$}
\ENDFOR
\STATE{Compute $\mathbf{C}_{l,t}^{P}$ as the average over all patch features (transformer features ${\mathbf{P}}_{l,t}^{n}$ for all selected patches and pseudo features $\widetilde{\mathbf{P}}_{l,t}^{n}$ for all unselected patches). In addition, average pooling all patch features to get $2\times 4$ local tokens $\{\mathbf{L}_{l,t}^{j}\}_{j=1}^{8};$
}
\STATE{$\{\mathbf{P}_{l+1,t}^{n}\}_{n=1}^{\hat{N}_{t}}={\rm MSA}_{l}(\{\mathbf{P}_{l,t}^{n}\}_{n=1}^{\hat{N}_{t}}||\mathbf{C}_{l,t}^{P}||\{\mathbf{L}_{l,t}^{j}\}_{j=1}^{8})$.}
\ENDIF
\ENDFOR
\STATE{\textbf{Return:} $\{\mathbf{I}_{l+1}^{i}\}_{i=1}^{N}$, $\{\{\mathbf{P}_{l+1,t}^{n}\}_{n=1}^{\hat{N}_{t}}\}_{t=1}^{T-1}$}
\end{algorithmic}
\end{algorithm}

\subsubsection{Transformer Structure }\label{text_trans}The pseudo context $\mathbf{C}_{l,t}^{P}$ obtained through dynamic routing is subsequently used to build our patch-spares transformer (PSFormer). Specifically, in the $l$-th layer of PSFormer, we denote the input features of all selected patches in the $t$-th P-frame as $\{\mathbf{P}_{l,t}^{n}\}_{n=1}^{\hat{N}_{t}}$. We append $\mathbf{C}_{l,t}^{P}$ as a token to $\{\mathbf{P}_{l,t}^{j}\}_{j=1}^{\hat{N}_{t}}$, and the resulted $[\mathbf{P}_{l,t}^{1}, \mathbf{P}_{l,t}^{2}, ..., \mathbf{P}_{l,t}^{\hat{N}_{t}},\mathbf{C}_{l,t}^{P}]$ is input into a multi-head self attention layer as in ViT. We observe that when patch-wise warping is used, the generated features not only encompass global context but also include the pseudo feature $\widetilde{\mathbf{P}}_{l,t}^{n}$ for each unselected patch. To fully and efficiently utilize this fine-grained information, we employ average pooling to generate $2\times 4$ tokens from these pseudo features. These tokens are also appended to $\{\mathbf{P}_{l,t}^{n}\}_{n=1}^{\hat{N}_{t}}$ as attention inputs. Consequently, PSFormer is constituted by successively stacking a total of $L$ layers with the aforementioned structure. In Algorithm \ref{alg_2}, we present the detailed procedures for each layer of our patch-sparse transformer. 

After the last layer of the patch-sparse transformer, we employ a patch-wise warping process to generate pseudo features for all patches in the P-frames that have not been selected. Subsequently, features for all patches in all frames, including transformer output features for selected patches and pseudo features for unselected patches, are averaged to obtain the final output of the patch-sparse transformer. This output feature is utilized for person re-identification. Note that our method can be applied to most of the existing video ReID models by replacing their used backbone network with our method that includes a patch selection stage followed by a patch-sparse transformer. These existing methods typically design modules to further capture more useful features, such as temporal information \cite{wang2021pyramid, yang2020spatial, bai2022salient} like the walking pattern of pedestrians, based on the features extracted from the backbone. In such cases, the output of the patch-sparse transformer can be used as the backbone feature for further processing.

\section{Loss and Training} \label{loss_text}
Following previous methods, we train our model using a combination of cross-entropy loss and hard triplet loss. In addition, we further propose an error-constraint loss function to ensure the effectiveness of our error measurement used in the dynamic routing mechanism. Specifically, in each transformer layer $l$, we randomly sample $S$ numbers from the interval [0, 1] and sort them ascendingly, yielding $\{\alpha_{i}\}_{i=1}^{S}$ with $\alpha_{S}>\alpha_{S-1}> ... >\alpha_{1}$. Each $\alpha_{i}$ is used as a weight to generate a noisy I-frame global context $\widetilde{\mathbf{C}}^{I}_{l,i}=(1-\alpha_{i})\mathbf{C}^{I}_{l}+\alpha_{i}\mathcal{N}$, where $\mathcal{N}$ represents a standard Gaussian noise. Subsequently, using the global-level warping, we generate an evolution feature from $\widetilde{\mathbf{C}}^{I}_{l,i}$ and ${\mathbf{C}}^{I}_{l-1}$, which is then used to obtain a reconstructed pseudo $\widetilde{\mathbf{C}}_{l}^{I}$ by warping from ${\mathbf{C}}^{I}_{l-1}$. We calculate the cosine distance $c_{l,i}$ between $\widetilde{\mathbf{C}}_{l,i}^{I}$ and the actual I-frame global context $\mathbf{C}_{l}^{I}$. Based on the intuition that a higher $\alpha_{i}$ can lead to a noisier $\widetilde{\mathbf{C}}^{I}_{l,i}$ and finally result in a higher distance $c_{l,i}$, we formulate the following constraint:
\begin{equation} \label{noise_loss}
    \mathcal{L}_{error}^{l} = \sum_{i=1}^{S-1}\sum_{j=i+1}^{S}{\rm max}\left(0,c_{l,i} - c_{l,j}\right).
\end{equation}
Using $\mathcal{L}_{error}$, we constrain the distance between the reconstructed I-frame features and the actual features to be positively correlated with the error amount. This constraint ensures that our error measurement method used in the dynamic routing mechanism can accurately reflect the accumulation condition of errors.

Using these losses, we adopt a 2-stage training process to optimize our method. In the first stage, we exclude the patch selection, feature warping, and dynamic routing components, using all patches from all frames as input to train the transformer layers in the patch-sparse transformer for 100 epochs. The loss function for this stage is the sum of cross-entropy loss $\mathcal{L}_{cent}$ and hard triplet loss $\mathcal{L}_{tri}$, i.e, $\mathcal{L}_{cent}+\mathcal{L}_{tri}$. In the second stage, we further train the complete model for another 100 epochs. The loss function for this stage is the sum of cross-entropy loss $\mathcal{L}_{cent}$, hard triplet loss $\mathcal{L}_{tri}$ and error-constraint loss $\mathcal{L}_{error}$, i.e, $\mathcal{L}_{cent}+\mathcal{L}_{tri}+\mathcal{L}_{error}$. Note that the training in the first stage is crucial. It allows the model to encounter more complete ReID images during training, which enables the model to learn better abilities to extract ReID-useful person features. We will show in experimental sections to demonstrate the effectiveness of our multi-stage training mechanism.

\section{Experiments}
\subsection{Implementation Details}
We use Adam as the optimizer with a momentum of 0.0005. The initial learning rate is set to 0.0005, which decays by 0.1 for every 40 epochs (overall 200 epochs). $s$ in Sec.\ref{text_dynamic} is set to 0.5. $S$ in Eq.\ref{noise_loss} is set to 4. We follow previous methods by using the restricted random sampling (RRS) strategy to sample 8 frames to generate each video clip, and the GOP that helps to perform the patch selection and patch-wise warping are obtained from this generated video clip after sampling. The frames are resized to $128\times 256$. We apply the commonly-used data augmentation strategies for training, including left-right flipping and random erasing. In PSFormer, the attention blocks follow the same setting as ViT with the same number of feature dimensions. We test and report the results of our method with 2 scales: Ours-small (Ours-S) with 8 layers and Ours-base (Ours-B) with 12 layers. We conduct experiments on the NVIDIA V100 GPUs.

\begin{table*}[t]
    \centering
    \caption{Comparison results when using five different networks (ResNet50, ViT-B, Ours-B, ViT-S, Ours-S) as the backbones for existing video-based ReID methods.}
    \setlength\tabcolsep{11pt}
    \begin{adjustbox}{width=2\columnwidth,center}
    \renewcommand{\arraystretch}{0.95}
    \begin{tabular}{l|c|c|c|cc|cc|c|c}
    \toprule
    & & & & \multicolumn{2}{c|}{\textbf{MARS}} & \multicolumn{2}{c|}{\textbf{LS-VID}} & \multicolumn{1}{c|}{\textbf{iLiDS-VID}} & \multicolumn{1}{c}{\textbf{PRID-2011}} \\
    Method & Backbone & GMACs & ms/video & mAP & rank-1 & mAP & rank-1 & rank-1 & rank-1 \\
    \midrule
    ResNet50 & ResNet50 & 32.7 & 94 & 81.0 & 88.1 & 68.5 & 80.4 & 83.3 & 90.2 \\
    ViT-B & ViT-B & 88.9 & 272 & 86.7 & 90.0 & 78.3 & 87.4 & 90.0 & 93.8 \\
    Ours-B & \cellcolor{mygray}Ours-B & \cellcolor{mygray}23.5 & \cellcolor{mygray}78 &  \cellcolor{mygray}86.1 & \cellcolor{mygray}89.5 & \cellcolor{mygray}77.7 & \cellcolor{mygray}86.9 & \cellcolor{mygray}89.3 & \cellcolor{mygray}93.4\\
    ViT-S & ViT-S & 51.1 & 144 & 85.4 & 89.2 & 77.0 & 85.8 & 88.5 & 93.1 \\
    Ours-S & \cellcolor{mygray}Ours-S & \cellcolor{mygray}14.4 & \cellcolor{mygray}45 &  \cellcolor{mygray}84.8 & \cellcolor{mygray}89.1 & \cellcolor{mygray}76.5 & \cellcolor{mygray}85.4 & \cellcolor{mygray}88.0 & \cellcolor{mygray}92.8\\
    \midrule
    \multirow{5}*{MGH \cite{yan2020learning}} & ResNet50 & 33.5 & 101 & 85.8 & 90.0 & 73.7 & 84.8 & 85.6 & 94.8 \\
     & ViT-B & 89.7 & 280 & 88.0 & 91.0 & 79.5 & 88.8 & 91.6 & 96.2 \\
     & \cellcolor{mygray}Ours-B & \cellcolor{mygray}24.2 & \cellcolor{mygray}85 & \cellcolor{mygray}87.6 & \cellcolor{mygray}90.8 & \cellcolor{mygray}79.1 & \cellcolor{mygray}88.5 & \cellcolor{mygray}91.3 & \cellcolor{mygray}96.0 \\
     & ViT-S & 52.0 & 153 & 87.1 & 90.6 & 78.7 & 88.0 & 91.0 & 95.7 \\
     & \cellcolor{mygray}Ours-S & \cellcolor{mygray}15.2 & \cellcolor{mygray}54 & \cellcolor{mygray}86.7 & \cellcolor{mygray}90.3 & \cellcolor{mygray}78.4 & \cellcolor{mygray}87.8 & \cellcolor{mygray}90.8 & \cellcolor{mygray}95.6 \\
     \midrule
     \multirow{5}*{SINet \cite{bai2022salient}} & ResNet50 & 37.7 & 115 & 86.3 & 91.2 & 79.6 & 87.6 & 92.7 & 96.5 \\
     & ViT-B & 93.9 & 297 & 88.5 & 92.2 & 81.8 & 89.7 & 94.0 & 97.6 \\
     & \cellcolor{mygray}Ours-B & \cellcolor{mygray}28.6 & \cellcolor{mygray}100 & \cellcolor{mygray}88.2 & \cellcolor{mygray}92.1 & \cellcolor{mygray}81.7 & \cellcolor{mygray}89.5 & \cellcolor{mygray}93.7 & \cellcolor{mygray}97.6 \\
     & ViT-S & 56.1 & 170 & 87.7 & 91.7 & 81.0 & 89.0 & 93.5 & 97.2 \\
     & \cellcolor{mygray}Ours-S & \cellcolor{mygray}19.4 & \cellcolor{mygray}70 & \cellcolor{mygray}87.5 & \cellcolor{mygray}91.6 & \cellcolor{mygray}80.8 & \cellcolor{mygray}88.8 & \cellcolor{mygray}93.3 & \cellcolor{mygray}97.1 \\
     \midrule
     \multirow{5}*{GRL \cite{liu2021watching}} & ResNet50 & 34.8 & 105 & 84.8 & 91.0 & 72.1 & 83.0 & 90.4 & 96.2 \\
     & ViT-B & 91.0 & 285 & 87.9 & 91.9 & 79.1 & 88.5 & 92.9 & 97.2 \\
     & \cellcolor{mygray}Ours-B & \cellcolor{mygray}25.6 & \cellcolor{mygray}88 & \cellcolor{mygray}87.5 & \cellcolor{mygray}91.7 & \cellcolor{mygray}78.7 & \cellcolor{mygray}88.4 & \cellcolor{mygray}92.6 & \cellcolor{mygray}97.2 \\
     & ViT-S & 53.3 & 159 & 87.0 & 91.5 & 78.0 & 87.7 & 92.0 & 97.0 \\
     & \cellcolor{mygray}Ours-S & \cellcolor{mygray}16.6 & \cellcolor{mygray}59 & \cellcolor{mygray}86.7 & \cellcolor{mygray}91.5 & \cellcolor{mygray}77.6 & \cellcolor{mygray}87.3 & \cellcolor{mygray}91.8 & \cellcolor{mygray}96.9 \\
     \midrule
     \multirow{5}*{STMN\cite{eom2021video}} & ResNet50 & 36.4 & 110 & 84.5 & 90.5 & 69.5 & 82.1 & 91.5 & 95.5 \\
     & ViT-B & 92.6 & 290 & 87.5 & 91.4 & 79.0 & 88.7 & 93.5 & 97.1 \\
     & \cellcolor{mygray}Ours-B & \cellcolor{mygray}27.2 & \cellcolor{mygray}94 & \cellcolor{mygray}87.3 & \cellcolor{mygray}91.4 & \cellcolor{mygray}78.6 & \cellcolor{mygray}88.4 & \cellcolor{mygray}93.4 & \cellcolor{mygray}97.0 \\
     & ViT-S & 92.6 & 165 & 86.8 & 91.1 & 78.0 & 87.8 & 93.0 & 96.6 \\
     & \cellcolor{mygray}Ours-S & \cellcolor{mygray}17.2 & \cellcolor{mygray}64 & \cellcolor{mygray}86.6 & \cellcolor{mygray}90.9 & \cellcolor{mygray}77.7 & \cellcolor{mygray}87.5 & \cellcolor{mygray}92.9 & \cellcolor{mygray}96.4 \\
     \bottomrule
    \end{tabular}
    \end{adjustbox}
    \label{main_compare}
\end{table*}

\subsection{Main Results} \label{comparision_text}
\subsubsection{Performance When Used as the Backbone for Existing Methods} Table \ref{main_compare} presents the performance and computational costs when using our method as the backbone for existing methods across four datasets: MARS, LS-VID, iLiDS-VID, and PRID-2011. Specifically, we select four existing video-based ReID models, including MGH \cite{yan2020learning}, SINet \cite{bai2022salient}, GRL \cite{liu2021watching}, and STMN\cite{eom2021video}, and replace their original backbone networks with the method proposed by us. We use two metrics to assess the model's performance: mean Average Precision (mAP) and rank-1, and two metrics to measure the computational cost of each method, including multiply–accumulate operations (MACs), which reflects the theoretical computation of the model; and milliseconds per video (ms/video), which calculates the total time for processing the whole 8 frames of each video on a realistic hardware platform (Tesla V100 in our experiments). The ms/video result is calculated as the average time for all test videos across all four datasets. Note that the reported computational cost of our method is the sum of the computational costs for both the patch selection stage and the patch-sparse transformer stage. In addition to our proposed method, we also test the method that uses two other widely-used networks as the model's backbone, including ResNet50 that is used by most of the previous methods, and the vanilla ViT without using our efficiency improvement strategies. For the comprehensive evaluation of method effectiveness, we validate and compare ViT and our method under two different sizes: ViT-B and Ours-B with 12 layers, and ViT-S and Ours-S with 8 layers. Based on the experimental results shown in Table \ref{main_compare}, we find that when used in conjunction with various existing methods, our proposed backbone can consistently achieve highly competitive results with only minimal computational requirements. Specifically, our approach (Ours-B) largely decreases computational cost by 74\% compared to ViT-B and 28\% compared to ResNet50, but can achieve results that closely resemble ViT-B and outperform ResNet50 significantly (+\%5.1 mAP on mAP and +\%8.5 on LS-VID). Additionally, with the same number of transformer layers, Ours-S model also achieves results close to ViT-S while significantly reducing the computational cost. Moreover, compared to ViT-S, which has fewer layers, the Ours-B model, despite having more network layers, still requires less computational cost while achieving better performance. It is worth noting that due to the information loss associated with sparse input, our method exhibits a slight performance gap compared to ViT. However, given its only 26\% computational requirement, our approach offers significantly higher practical utility value in real-time applications. Moreover, in comparison to the most-frequently-used ResNet50, our method not only reduces computational demands but also substantially enhances performance. Therefore, our method can be used to replace ResNet50 as a plug-and-play backbone, which can be seamlessly integrated with any other method to achieve both better performance and higher efficiency.

\begin{table}[t]
    \centering
    \caption{Comparisons with other efficient transformer methods.}
    \begin{adjustbox}{width=1.0\columnwidth,center}
    \setlength\tabcolsep{6pt}
    \renewcommand{\arraystretch}{1.0}
    \begin{tabular}{l| c | c c|c c}
    \toprule
    & & \multicolumn{2}{c|}{\textbf{MARS}} & \multicolumn{2}{c}{\textbf{LS-VID}} \\
    Method & GMACs & mAP & rank-1 & mAP & rank-1\\
    \midrule
    ViT-B \cite{dosovitskiy2020image} & 88.9 & 86.7 & 90.0 & 78.3 & 87.4\\
    \midrule
    DynamicViT \cite{rao2021dynamicvit} & 35.5 & 83.8 & 88.4 & 74.8 & 84.7\\
    Evit \cite{liang2022not} & 35.0 & 84.2 & 88.6 & 75.6 & 85.3\\
    SPViT \cite{kong2022spvit} & 35.1 & 83.7 & 88.3 & 75.3 & 85.3\\
    dTPS \cite{wei2023joint} & 35.4 & 84.1 & 88.6 & 75.6 & 85.3 \\
    MViT \cite{wang2022efficient} & 31.9 & 84.8 & 89.0 & 76.6 & 85.9 \\
    DiffRate \cite{chen2023diffrate} & 34.5 & 84.5 & 88.9 & 76.4 & 85.7 \\
    \midrule
    \rowcolor{mygray} Ours-B & 23.5 & 86.1 & 89.5 & 77.7 & 86.9 \\
    \bottomrule
    \end{tabular}
    \end{adjustbox}
    \label{transformer_compare}
\end{table}

\begin{table}[t]
    \centering
    \setlength\tabcolsep{4.3pt}
    \caption{Comparison of using different efficient transformer methods as the backbones for existing video-based ReID methods.}
    \begin{adjustbox}{width=1\columnwidth,center}
    \renewcommand{\arraystretch}{1.0}
    \begin{tabular}{l|c|c|c c |c c}
    \toprule
    & & & \multicolumn{2}{c|}{\textbf{MARS}} & \multicolumn{2}{c}{\textbf{LS-VID}}\\
    Method & Backbone & GMACs & mAP & rank-1 & mAP & rank-1 \\
    \midrule
    \multirow{7}*{MGH \cite{yan2020learning}}
     & DynamicViT \cite{rao2021dynamicvit} & 36.2 & 85.8 & 89.7 & 76.5 & 85.5\\
     & Evit \cite{liang2022not} & 35.7 & 86.2 & 90.1 & 77.0 & 86.2\\
     & SPViT \cite{kong2022spvit} & 35.8 & 85.8 & 89.9 & 77.3 & 86.1 \\
     & dTPS \cite{wei2023joint} & 36.1 & 86.4 & 90.1 & 77.4 & 86.4\\
     & MViT \cite{wang2022efficient} & 32.6 & 86.7 & 90.3 & 78.1 & 86.9\\
      & DiffRate \cite{chen2023diffrate} & 34.7 & 86.4 & 90.1 & 77.9 & 86.8\\
     & \cellcolor{mygray}Ours-B & \cellcolor{mygray}24.2 & \cellcolor{mygray}87.6 & \cellcolor{mygray}90.8 & \cellcolor{mygray}79.1 & \cellcolor{mygray}88.5 \\
     \midrule
        \multirow{7}*{GRL \cite{liu2021watching}}
     & DynamicViT \cite{rao2021dynamicvit} & 37.6 & 84.7 & 90.7 & 76.1 & 86.0\\
     & Evit \cite{liang2022not} & 36.9 & 85.0 & 91.0 & 76.9 & 86.5\\
     & SPViT \cite{kong2022spvit} & 37.0 & 85.0 & 90.9 & 77.2 & 86.4 \\
     & dTPS \cite{wei2023joint} & 37.3 & 85.2 & 91.1 & 77.1 & 86.8\\
     & MViT \cite{wang2022efficient} & 33.8 & 85.6 & 91.2 & 78.1 & 87.6\\
      & DiffRate \cite{chen2023diffrate} & 35.9 & 85.5 & 91.2 & 77.7 & 87.3\\
     & \cellcolor{mygray}Ours-B & \cellcolor{mygray}25.6 & \cellcolor{mygray}87.5 & \cellcolor{mygray}91.7 & \cellcolor{mygray}78.7 & \cellcolor{mygray}88.4 \\
     \bottomrule
    \end{tabular}
    \end{adjustbox}
    \label{supp_compare}
\end{table}

\subsubsection{Comparison with Other Efficient Transformers}We further compare our approaches with other transformer methods that have the same objective of achieving high efficiency by using token pruning or selection techniques. To ensure fair comparisons, all these methods are evaluated using the same ViT-B network architecture and an equal number of layers. The results are presented in Table \ref{transformer_compare}, which includes both ReID performance and computational costs on the MARS and LS-VID datasets. The original ViT-B has the best performance, but it is very computationally intensive. Using token pruning or token selection techniques, the other methods can reduce computation while suffering from a slight loss in precision. This reduction in performance is primarily due to two reasons. First, the pruning or selection mechanisms of these methods are not specifically designed for video-based person ReID, so directly applying them to this task may not guarantee that the task-optimal patches are selected. Second, most of these methods focus only on selecting sparse tokens, without further considering how to address the loss of information resulting from pruning tokens. Compared to these methods, our method proposes to use ReID-task-tailored designs (patch novelty and patch semantics as selection-determinate features) for patch selection, so it can achieve the lowest computational costs while maintaining good performance by discarding redundant or unimportant patches to the greatest extent and preserving ReID-necessary patches as completely as possible. Additionally, we have innovatively designed a novel perception enhancement method for global information to address the issue of information loss when a lot of patches are discarded. By using these designs, we can still maintain good effectiveness while taking advantage of the reduced computational demand of sparse inputs. The results shown in Table \ref{transformer_compare} demonstrate that our method is both the most computationally efficient and performance-effective. Additionally, we also conduct performance comparisons when using these methods as the backbones for existing ReID methods. As shown in Table \ref{supp_compare}, in this case, our method still achieves the best performance and highest efficiency. These results further demonstrate that our method is an excellent choice to serve as a plug-and-play backbone for other video ReID methods to achieve high efficiency and performance.

\begin{table}[t]
    \centering
    \setlength\tabcolsep{8pt}
    \caption{Comparison with other state-of-the-art methods.}
    \begin{adjustbox}{width=1.0\columnwidth,center}
    \renewcommand{\arraystretch}{1.0}
    \begin{tabular}{l|l|c c |c c}
    \toprule
    & & \multicolumn{2}{c|}{\textbf{MARS}} & \multicolumn{2}{c}{\textbf{iLiDS-VID}}\\
    \midrule
    Method & Venue & mAP & rank-1 & rank-1 & rank-5 \\
    \midrule
    GRL \cite{liu2021watching} & CVPR2021 & 84.8 & 91.0 & 90.4 & 98.3\\
    CTL \cite{liu2021spatial} & CVPR2021 & 86.7 & 91.4 & 89.7 & 97.0\\
    STRF \cite{aich2021spatio} & ICCV2021 & 86.1 & 90.3 & 89.3 & -\\
    STMN \cite{eom2021video} & ICCV2021 & 84.5 & 90.5 & 91.5 & 98.5\\
    PSTA \cite{wang2021pyramid} & ICCV2021 & 85.8 & 91.5 & 91.5 & 98.1\\
    SINet \cite{bai2022salient} & CVPR2022 & 86.3 & 91.2 & 92.7 & 98.7\\
    CAVIT \cite{wu2022cavit} & ECCV2022 & 87.2 & 90.8 & 93.3 & 98.0\\
    DCCT \cite{liu2023deeply}& TNNLS2023 & 87.5 & 92.3 & 91.7 & 98.6\\
    MSTAT \cite{tang2022multi} & TMM2023 & 85.3 & 91.8 & 93.3 & 99.3\\
    FIDN \cite{liu2023frequency} & TIP2023 & 86.8 & 91.5 & 91.3 & 98.0\\
    \midrule
    \rowcolor{mygray} Ours & - & 88.2 & 92.1 & 93.7 & 99.5\\
     \bottomrule
    \end{tabular}
    \end{adjustbox}
    \label{comp_sota}
\end{table}

\subsubsection{Comparison with State-of-the-art Methods}We further compare our method with other state-of-the-art video-based ReID methods to demonstrate our superior performance. All the methods used for comparison are advanced works published in top-tier conferences or journals in the past three years, and the reported results for `ours' are obtained using SINet as the baseline with our method as its backbone. As shown in Table \ref{comp_sota}, across the four metrics on two datasets (MARS and iLiDS-VID), our method achieves the highest performance on three metrics (mAP for MARS, rank-1 and rank-5 for iLiDS-VID) and ranks second on the remaining one (rank-1 for MARS). These results highlight the high effectiveness and superiority of our method compared to the previous SOTA methods. Notably, almost all the compared methods use conventional CNNs or transformer networks as the backbone for feature extraction, while our method significantly reduces the backbone's computational demands through the innovatively proposed patch selection mechanism, which makes the network more efficient.

\subsection{Ablation Study} \label{abaltion}
We further perform ablation study to verify the effectiveness of our designs. Experiments in this section are conducted on MARS dataset based on the Ours-B model.

\begin{table}[t]
    \centering
    \caption{Ablation results of patch selection.}
    \begin{adjustbox}{width=1.0\columnwidth,center}
    \renewcommand{\arraystretch}{1.0}
    \setlength\tabcolsep{5pt}
    \begin{tabular}{l | c c c}
    \toprule
    Method & mAP & rank-1 & GMACs \\
    \midrule
    \rowcolor{mygray}Ours & 86.1 & 89.5 & 23.5\\
    \midrule
    Ours w/o patch selection & 86.7 & 90.0 & 88.9\\
    Ours w/o preserving whole I-frame & 82.5 & 87.9 & 20.2\\
    Ours w/o using patch novelty $\mathbf{N}$ & 86.3 & 89.6 & 70.9\\
    Ours w/o using patch semantics $\mathbf{S}$ & 84.5 & 88.6 & 39.0\\
    Ours w/o enhancement for $\mathbf{S}$ & 85.6 & 89.0 & 31.5\\
    Ours w/o frame-progressive selection ($\mathbf{R}$) & 86.1 & 89.5 & 33.4\\
     \bottomrule
    \end{tabular}
    \end{adjustbox}
    \label{ablation_selection}
\end{table}

\subsubsection{Ablation of Patch Selection}Table \ref{ablation_selection} presents the evaluation of our proposed patch selection mechanism. By removing patch selection and using all patches from a video for feature extraction, the computational cost is greatly increased by 278\%. This result shows the significant role of our patch selection mechanism in reducing computation costs and achieving a high-efficient model. We further validate the contributions of various designs and components within our patch selection mechanism, including (1) the retention of the whole I-frame for feature extraction, (2) patch novelty $\mathbf{N}$ and patch semantics $\mathbf{S}$ as the selection-determinate features, (3) spectral decomposition methods used to enhance patch semantics $\mathbf{S}$, and (4) the frame-progressive strategy with the factor $\mathbf{R}$ as a selection-determinate feature. The results indicate that removing any of these designs or components would result in either a significant decrease in model performance or a substantial increase in computational requirements. These results demonstrate that all designs in our mechanism can contribute to selecting patches that are more valuable and non-redundant, thus helping to achieve both high performance and reduced computational cost.

\subsubsection{Ablation of PSFormer}Table \ref{ablation_psformer} presents an evaluation of different design choices in our PSFormer framework. (1) By excluding the use of context in each transformer layer, the mAP decreases by 3.2. This highlights the importance of incorporating global context during the feature extraction process, which helps to mitigate the problem of information loss caused by the sparse inputs of the network, thereby enhancing the representation ability of the features extracted by the transformer. (2) To generate the pseudo P-frame context feature, we employ a dynamic routing mechanism to select a feature warping method in each transformer layer. Without this mechanism and solely using global-level warping in all layers, the mAP decreases by 2.2. This is because global-level warping only utilizes a generalized global information for further processing, thus the obtained pseudo context lacks fine-grained detail features, which leads to the reduced accuracy. On the other hand, solely using patch-wise warping achieves slightly better performance than our method by utilizing more fine-grained information, but it requires a significantly higher computation cost (+71\%) due to the patch-level operations and feature fusion mechanism. As demonstrated by the experimental results, by dynamically and adaptively selecting between two complementary warping methods using the proposed dynamic routing mechanism, our method can achieve the optimal balance between low computation and high effectiveness.

\begin{table}[t]
    \centering
     \caption{Ablation results of PSFormer.}
    \begin{adjustbox}{width=1.0\columnwidth,center}
    \renewcommand{\arraystretch}{1.0}
    \setlength\tabcolsep{5pt}
    \begin{tabular}{l | c c c}
    \toprule
    Method & mAP & rank-1 & GMACs \\
    \midrule
    \rowcolor{mygray}Ours & 86.1 & 89.5 & 23.5\\
    \midrule
    Ours w/o using context in each layer & 82.9 & 88.2 & 18.5\\
    Ours solely using global-level warping & 83.9 & 88.5 & 19.1\\
    Ours solely using patch-wise warping & 86.4 & 89.7 & 36.5\\
     \bottomrule
    \end{tabular}
    \end{adjustbox}
    \label{ablation_psformer}
\end{table}

\begin{table}[t]
    \centering
    \caption{Different methods to select the warping method in each network layer.}
    \begin{adjustbox}{width=1.0\columnwidth,center}
    \renewcommand{\arraystretch}{1.0}
    \setlength\tabcolsep{5pt}
    \begin{tabular}{l | c c c}
    \toprule
    Method & mAP & rank-1 & GMACs \\
    \midrule
    \rowcolor{mygray}Noise-conditioned Gate (Our method) & 86.1 & 89.5 & 23.5\\
    \midrule
    Randomly choosing in each layer & 84.5 & 88.5 & 24.0\\
    Using patch-wise warping every 4 layers & 84.9 & 88.7 & 23.9\\
     \bottomrule
    \end{tabular}
    \end{adjustbox}
    \label{ablation_gate}
\end{table}

\subsubsection{Effectiveness of Noise-conditioned Gate}
The proposed dynamic routing mechanism in our method employs a noise-conditioned gate to select between the global-level warping and patch-wise warping in each network layer. To verify the effectiveness of this gate, we compare it with two other strategies and present the results in Table \ref{ablation_gate}. Specifically, the first comparison strategy uses a random warping method in each layer instead of using the proposed selection method, which decreases the mAP by 1.6. The second comparison strategy is to use patch-wise warping every four layers. This method also decreases the mAP by 1.2. These results demonstrate that, compared to random or manually designed methods, the proposed noise-conditioned gate can help the network to select the more appropriate warping method in each layer automatically and self-adaptively, thus achieving a better balance between computation and performance.

\subsubsection{Dynamic Adjustment for Computation-Performance Balance}
As introduced detailedly in Sec.\ref{text_dynamic}, in the proposed dynamic routing mechanism, we introduce a hyper-parameter $s$ to control the activation of the noise-conditioned gate. We find that by setting $s$ to different values, we can dynamically adjust the balance between model computation and ReID performance based on user requirements. Specifically, if higher performance is the user's focus, the value of $s$ can be decreased, which results in more network layers using the more accurate but computationally demanding patch-wise warping, thus leading to better performance but more computation cost. Conversely, if a faster processing speed is desired, the value of $s$ can be increased to achieve faster processing but with a slight sacrifice in accuracy by using global-level warping more frequently. 
In Table. \ref{ablation_s}, we show how the model computation and performance change as $s$ increases, which demonstrates the high flexibility and controllability of our approach that is valuable in real-world applications. 

\begin{table}[t]
    \centering
    \caption{Performance and computation for different thresholds $s$.}
    \begin{adjustbox}{width=1.0\columnwidth,center}
    \setlength\tabcolsep{8pt}
    \begin{tabular}{l | c c c c c c c}
    \toprule
    Threshold $s$ & 0.4 & 0.5 & 0.6 & 0.7 & 0.8 & 0.9 \\
    \midrule
    mAP & 86.3 & 86.1 & 85.5 & 85.0 & 84.6 & 84.2\\
    rank-1 & 89.5 & 89.5 & 89.1 & 88.9 & 88.7 & 88.5\\
    GMACs & 27.1 & 23.5 & 21.9 & 20.9 & 19.9 & 19.5\\
     \bottomrule
    \end{tabular}
    \end{adjustbox}
    \label{ablation_s}
\end{table}

\begin{table}[t]
    \centering
    \caption{Ablation results of loss functions and training methods. }
    \begin{adjustbox}{width=1.0\columnwidth,center}
    \renewcommand{\arraystretch}{1.0}
    \setlength\tabcolsep{13pt}
    \begin{tabular}{l | c c}
    \toprule
    Method & mAP & rank-1\\
    \midrule
    \rowcolor{mygray} Ours & 86.1 & 89.5\\
    \midrule
    Ours w/o $\mathcal{L}_{cent}$ & 79.2 & 85.8\\
    Ours w/o $\mathcal{L}_{tri}$ & 82.9 & 87.9\\
    Ours w/o $\mathcal{L}_{error}$ & 84.9 & 88.5\\
    \midrule
    Ours w/o training in the first stage & 83.3 & 88.0\\
     \bottomrule
    \end{tabular}
    \end{adjustbox}
    \label{ablation_loss}
\end{table}

\subsubsection{Evaluation of Loss and Training Methods} \label{text_eval_loss_train}
\begin{figure}[t]
    \centering
    \includegraphics[width=0.9\linewidth]{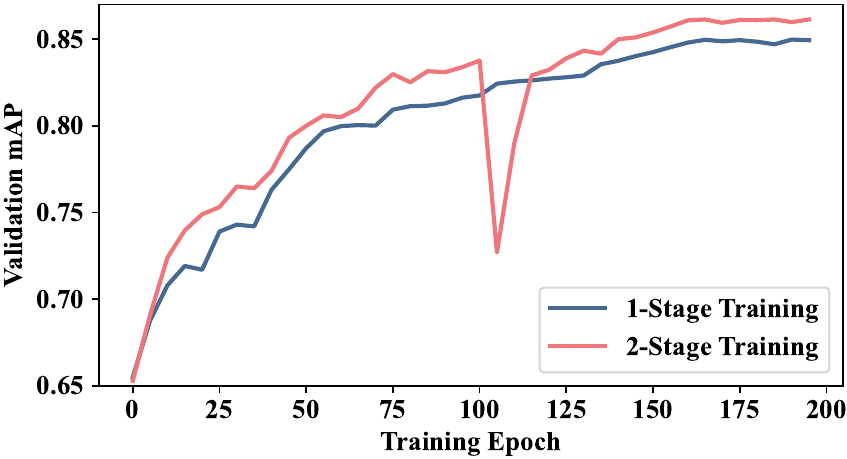}
    \caption{Convergence curves on the MARS dataset.}
    \label{curve}
\end{figure}

We further evaluate the effectiveness of the proposed loss functions and two-stage training methods. The results are presented in Table \ref{ablation_loss}. As introduced in Sec.\ref{loss_text}, the loss function used to optimize our method is the sum of three components: cross-entropy loss $\mathcal{L}_{cent}$, hard triplet loss $\mathcal{L}_{tri}$, and error-constraint loss $\mathcal{L}_{error}$. As shown in Table \ref{ablation_loss}, removing any one of these three components will significantly reduce model performance, which demonstrates the rationality of using all three functions together. Notably, the error-constraint loss is innovatively proposed by us and is specifically designed for the dynamic routing mechanism in our PSFormer. It ensures that the error measurement method used in the mechanism can accurately reflect the error accumulation condition. The drop in performance when removing $\mathcal{L}_{error}$ demonstrates the importance of this newly proposed loss function. Furthermore, we validate the two-stage training method used for optimizing our model. As discussed in Sec \ref{loss_text}, in the first stage, we use all patches from all frames as input to only train the transformer layers in the PSFormer. In the second stage, we further train the complete model with the patch selection modules. The training in the first stage is crucial since it allows the model to encounter more complete ReID images, which enables the model to learn better abilities to extract ReID-useful person features. Thus, compared to the training method only with the second stage, our two-stage training method can increase the mAP by 2.8.\footnote{For fair comparisons, we use the same total number of epochs for training the 1-stage method and 2-stage method.} In addition, we present the training curves to further demonstrate the advantages of our 2-stage training method. As shown in Figure \ref{curve}, compared to training for 200 epochs using only the second stage, training with the first stage for 100 epochs followed by the second stage for another 100 epochs in our method achieves better convergence. Note that the training curve of our 2-stage method shows a sudden change at the 100th epoch, as this is when the second stage begins, and additional modules are introduced (including the patch selection modules, as well as the feature warping and dynamic routing components in the patch-sparse transformer). However, the model quickly converges afterward, ultimately achieving better performance than the single-stage method. These results demonstrate the rationality and significant effectiveness of our proposed training methods.

\begin{figure}
    \centering
    \includegraphics[width=1\linewidth]{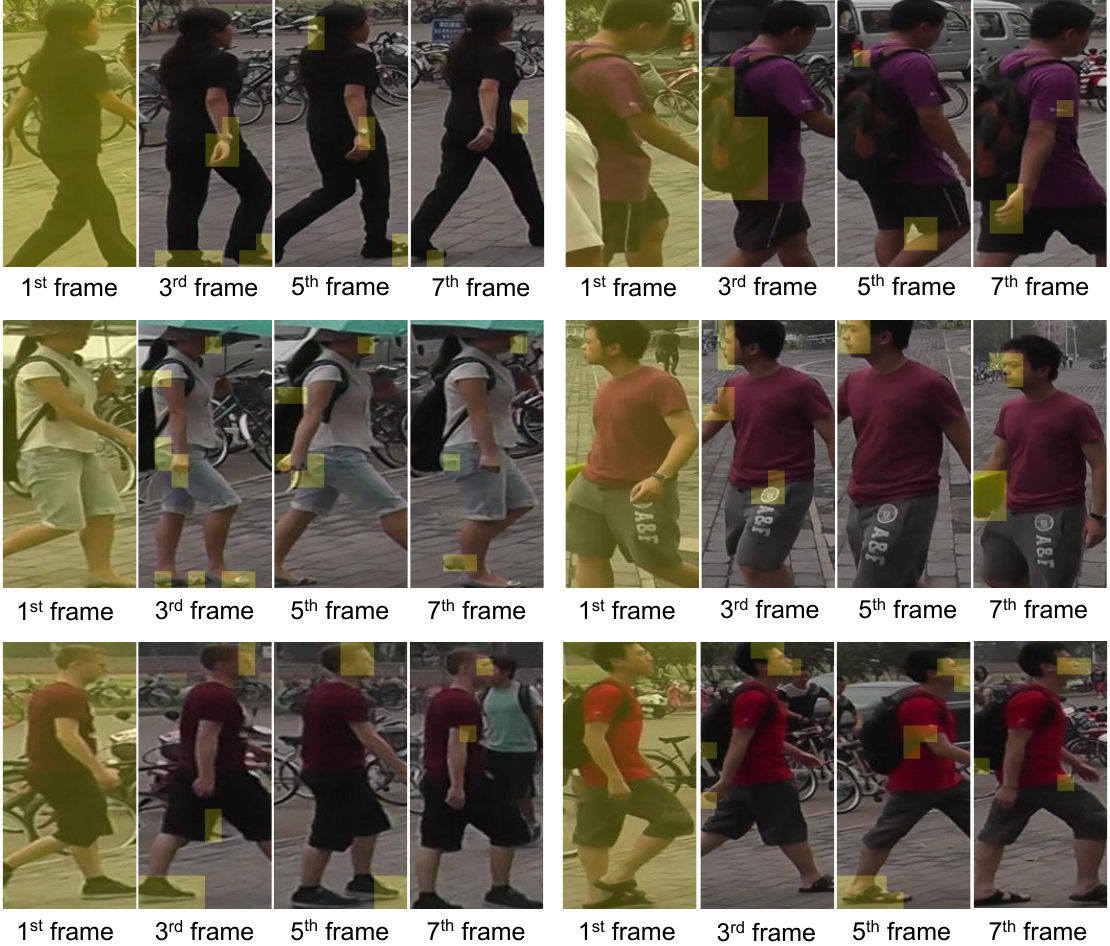}
    \caption{\textbf{Visualization of selected patches }for the 1st, 3rd, 5th, and 7th frames in a video clip. The selected patches are marked by yellow masks. (Best viewed in color)}
    \label{visual}
\end{figure}

\subsection{Further Analysis}

\subsubsection{Visualizations of Selected Patches}
To gain a deeper and more intuitive understanding of our patch selection method, in Figure \ref{visual}, we present some examples of the selected patches across different frames of a video clip. Specifically, we sample the 1st, 3rd, 5th, and 7th frames in a video clip for presentation, and the selected patches in these frames are highlighted in yellow. A video clip's first frame is regarded as the I-frame, and in our method, all patches within it are selected to extract global information. While for the following P-frames, we observe that only a few key patches are chosen while most redundant patches repeating from previous frames or located in background regions are effectively pruned. Interestingly, we find that a small number of patches that repeat across multiple frames are still consistently reselected. A reason behind this phenomenon could be that these patches are typically situated in crucial areas for ReID such as the human face, so the repetitive feature extraction for them can effectively emphasize and enhance the ReID-important information. Furthermore, as shown in the last row of Figure \ref{visual}, we observe that in a challenging condition when the target person is occluded by other persons in later P-frames of a video clip, our proposed method is still capable of making the correct patch selection decisions by excluding and not choosing the regions where these occluded persons are located. This is because our patch selection method uses patch semantics information as a selection-determinant feature, which is obtained by passing the video clip through a shallow 3D-CNN network. As such, it contains temporal information that can help differentiate the target person from the occluded persons in the video. These results demonstrate that our method can effectively select the most appropriate patches even in complex videos, thereby greatly reducing the computational cost of the subsequent transformer to realize a more efficient video ReID backbone.
\begin{figure}
    \centering
    \includegraphics[width=1\linewidth]{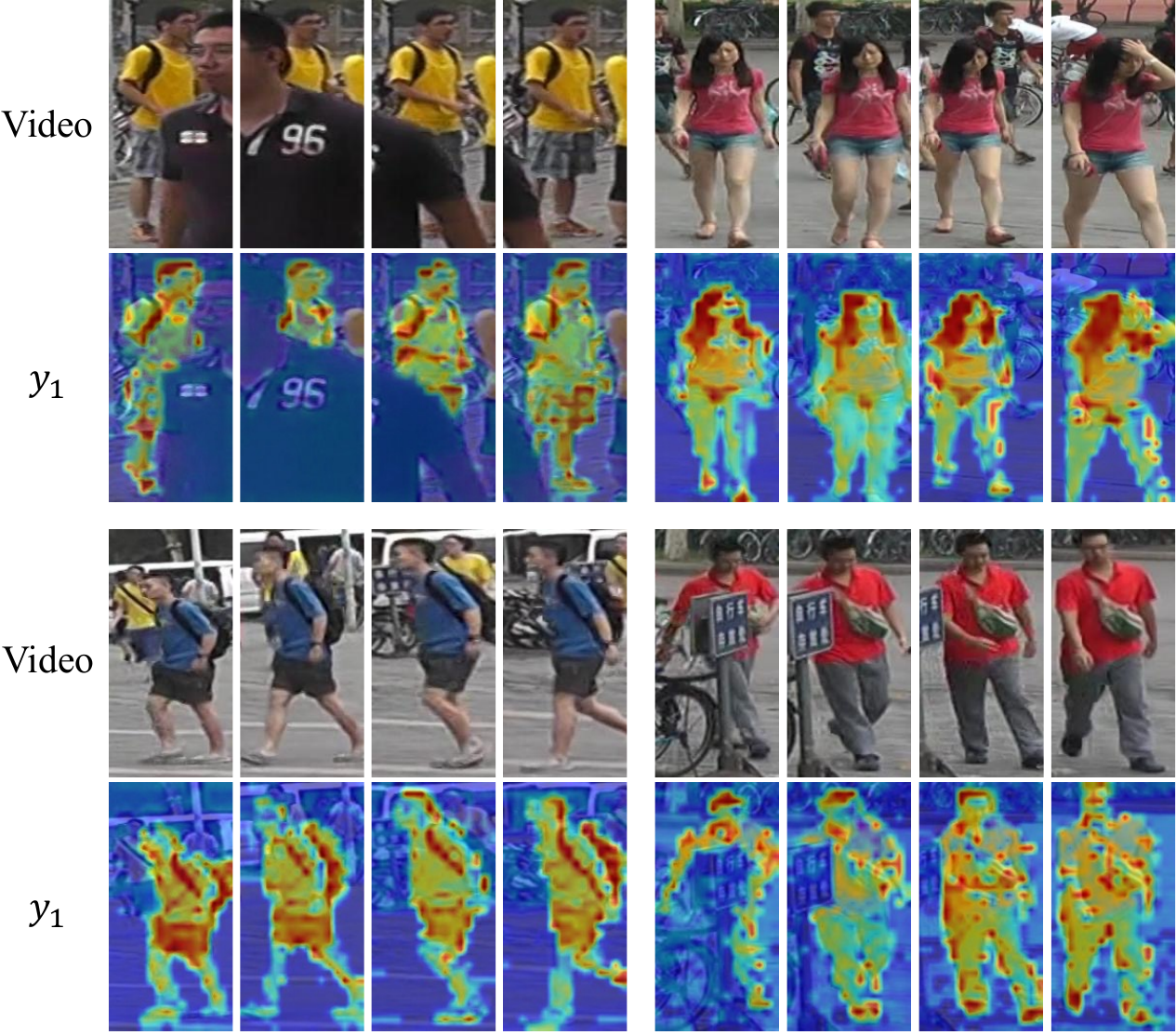}
    \caption{Visualization of the eigenvector $y_{1}$ associated with the smallest nonzero eigenvalue obtained through the spectral decomposition method illustrated in Sec.\ref{text_spectral}}
    \label{visual_spectral}
\end{figure}

\subsubsection{Effectiveness of Spectral Decomposition}\label{text_effect_spectral} As detailed in Sec.\ref{text_selection_feature}, to select only foreground target person regions for deep feature extraction, we employ spectral decomposition to compute an eigenvector associated with the smallest nonzero eigenvalue. This eigenvector can represent the most prominent object in a scene, which, in a ReID image, is typically the target person. To demonstrate this, we further conduct the following visual verification and quantitative validations: (1) \textbf{\textit{Visual verification}.} As shown in Figure \ref{visual_spectral}, we visualize the eigenvectors $y_{1}$ associated with the smallest nonzero eigenvalue from several ReID video frames. Specifically, a heat map is generated based on the value of $y_{1}$ in each patch for visualization. It can be seen that these eigenvectors effectively capture the region of the target person, even in frames where the background contains distracting people or objects. This benefits from the temporal information contained in $F_{t}$, which serves as the input feature for spectral decomposition and can help accurately distinguish the target person from background objects. These visualization results demonstrate the effectiveness of our method. (2) \textbf{\textit{Quantitative validation}.} With the assistance of the Segment Anything Model \cite{ravi2024sam}, we obtain pixel-level ground truth masks for the region of the target person in each video frame from the MARS validation set. We then apply our method to compute the eigenvector associated with the smallest nonzero eigenvalue for each frame, generate a binary mask using a threshold of 0, and calculate its mIoU with the corresponding ground truth mask. Across the entire dataset, our method achieves a high mIoU of 74.1\%. This result reveals that the prominent object represented by these eigenvectors overlaps highly with the target person, further demonstrating the high effectiveness of our method.

\begin{table}[t]
    \centering
    \setlength\tabcolsep{5pt}
    \caption{Ablation study for training epochs in different stages.}
    \begin{adjustbox}{width=0.78\columnwidth,center}
    \renewcommand{\arraystretch}{0.95}
    \begin{tabular}{c c|c c}
    \toprule
    {\scriptsize Stage1 Epoch} & {\scriptsize Stage2 Epoch} & {\scriptsize MARS mAP} & {\scriptsize MARS rank-1} \\
    \midrule
    70 & 130 & 86.00 & 89.34\\
    80 & 120 & 86.12 & 89.50\\
    90 & 110 & 86.18 & 89.50\\
    100 & 100 & 86.13 & 89.53\\
    110 & 90 & 86.08 & 89.46\\
    120 & 80 & 86.11 & 89.43\\
    130 & 70 & 86.03 & 89.39\\
     \bottomrule
    \end{tabular}
    \end{adjustbox}
    \label{table_ablation_hyper}
\end{table}

\begin{figure}[t]
    \centering
    \includegraphics[width=0.75\linewidth]{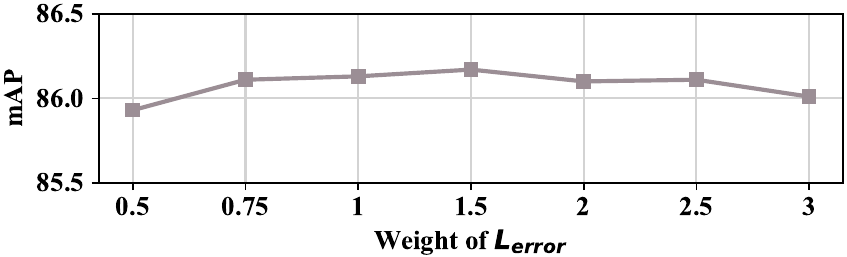}
    \caption{Ablation study for the weight of error-constraint loss $\mathcal{L}_{error}$ on MARS dataset.}
    \label{ablation_weight}
\end{figure}

\subsubsection{Hyperparameter Analysis} We further validate the robustness of our method to two key hyperparameter settings: (1) the number of training epochs in each stage, and (2) the weight of each loss function. To be specific, we keep the total number of training epochs at 200, vary the number of training epochs in the first stage to range from 70 to 130, and adjust the number of epochs in the second stage accordingly. As shown by the results presented in Table \ref{table_ablation_hyper}, our method consistently achieves stable and excellent performance across all evaluated settings on the MARS dataset. We then evaluate the performance of our method under varying weights for each loss function. Specifically, in the second training stage, our approach employs a weighted sum of three loss functions for optimization: cross-entropy loss $\mathcal{L}_{cent}$, hard triplet loss $\mathcal{L}_{tri}$, and error-constraint loss $\mathcal{L}_{error}$. For $\mathcal{L}_{cent}$ and $\mathcal{L}_{tri}$, we directly follow most previous works \cite{bai2022salient, yan2020learning, wang2021pyramid} by setting their weights to 1. Therefore, we primarily focus on the weight $W_{error}$ for our newly proposed function $\mathcal{L}_{error}$. The results presented in Figure \ref{ablation_weight} indicate that our method maintains stable performance across a wide range of $W_{error}$ values from 0.5 to 3. These results suggest that our method is not sensitive to the hyperparameter settings, since it can consistently achieve stable and excellent performance across a broad spectrum of hyperparameter values, demonstrating its high effectiveness and significant robustness.

\begin{figure}[t]
    \centering
    \includegraphics[width=1\linewidth]{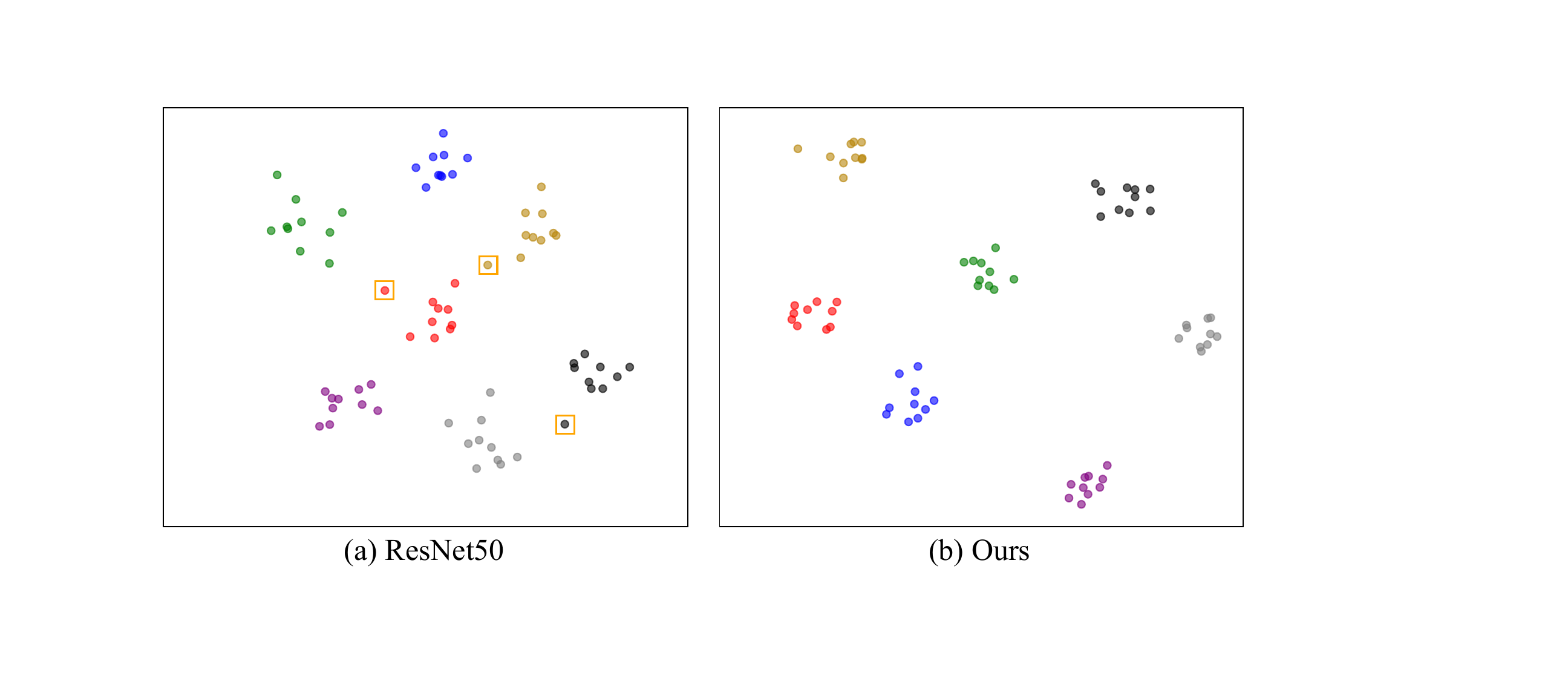}
    \caption{t-SNE visualization results of features from ResNet50 and our method. Circles marked by the orange bounding boxes refer to videos likely to be matched to the incorrect identities.}
    \label{tsne}
\end{figure}

\subsubsection{t-SNE Visualization Analysis} To further validate the effectiveness of our method, we provide the t-SNE visualizations of network features in Figure \ref{tsne}. In this visualization, each color represents a different identity, and each circle corresponds to features extracted from a video clip. It is observed that with the most-commonly-used ResNet50 as the backbone, features of different videos from the same identity exhibit significant intra-identity variation and are not tightly clustered. Moreover, the distances between distributions of different identities are relatively close, which may cause some videos at the distribution boundaries—such as the ones marked by orange bounding boxes—to be matched to the incorrect identities. In contrast, our method shows clear advantages: it significantly reduces intra-identity variation and enhances the distinctness of inter-identity boundaries. This improvement contributes to more accurate person re-identification, demonstrating the superior effectiveness of our approach compared to ResNet50, even with lower computational costs.

\section{Conclusion}
This paper proposes a new effective and efficient plug-and-play backbone for video person ReID. This backbone consists of two components: a patch selection mechanism to prune redundant patches for reduced computational load, and a patch-sparse transformer that achieves high-performance feature extraction enhanced by pseudo frame-global context. In comparison to the widely used ResNet50 backbone, our approach significantly reduces computational costs while effectively improving performance. We regard our method as a general and practical approach that can be utilized in real-world applications.

{
\bibliographystyle{ieee_fullname}
\bibliography{egbib}
}

\end{document}